\def\year{2021}\relax
\documentclass[letterpaper]{article} 
\usepackage{aaai21}  
\usepackage{times}  
\usepackage{helvet} 
\usepackage{courier}  
\usepackage[hyphens]{url}  
\usepackage{graphicx} 
\usepackage{natbib}  
\usepackage{caption} 
\usepackage[switch]{lineno}
\usepackage[export]{adjustbox}
\usepackage{amsmath}
\usepackage{amssymb}
\usepackage{mathtools}

\usepackage{comment}
\usepackage{bm}
\usepackage{mdframed}
\usepackage{tabularx}
\usepackage{booktabs}
\usepackage{multirow}
\usepackage{fancyhdr}
\usepackage{enumitem}
\usepackage{qtree}
\usepackage{newfloat}
\usepackage{fixmath}
\usepackage{subfig}
\usepackage[titletoc]{appendix}
\usepackage{soul}

\renewcommand{\vec}[1]{\mathbold{#1}}

\newcommand{\txt}{\mathbold{r}}
\newcommand{\txtFeat}{\mathbold{p}}

\newcommand{\txtReal}{\mathbold{r^+}}
\newcommand{\txtRealFeat}{\mathbold{p}^+}

\newcommand{\txtWrong}{\mathbold{r}^-}
\newcommand{\txtWrongFeat}{\mathbold{p}^-}

\newcommand{\img}{\mathbold{v}}
\newcommand{\imgFeat}{\mathbold{q}}

\newcommand{\imgReal}{\mathbold{v^+}}
\newcommand{\imgRealFeat}{\mathbold{q}^+}

\newcommand{\imgWrong}{\mathbold{v^-}}
\newcommand{\imgWrongFeat}{\mathbold{q}^-}

\DeclareMathOperator{\R}{\mathbb{R}}
\DeclareMathOperator{\TxtEnc}{F_p}
\DeclareMathOperator{\ImgEnc}{F_q}

\title{CHEF: Cross-modal Hierarchical Embeddings for Food Domain Retrieval}
\author{
    Hai X. Pham\textsuperscript{\rm 1},
    Ricardo Guerrero\textsuperscript{\rm 1},
    Jiatong Li\textsuperscript{\rm 2},
    Vladimir Pavlovic\textsuperscript{\rm 1}\textsuperscript{\rm 2} \\
}
\affiliations{
    \textsuperscript{\rm 1} Samsung AI Center, Cambridge \\
    \textsuperscript{\rm 2} Department of Computer Science, Rutgers University \\
    \{hai.xuanpham, r.guerrero, v.pavlovic\}@samsung.com, jl2312@rutgers.edu
}

\begin{document}
\pagenumbering{arabic}

\maketitle
\thispagestyle{empty}

\begin{abstract}
Despite the abundance of multi-modal data, such as image-text pairs, there has been little effort in understanding the individual entities and their different roles in the construction of these data instances. In this work, we endeavour to discover the entities and their corresponding importance in cooking recipes \textit{automatically} as a visual-linguistic association problem. More specifically, we introduce a novel cross-modal learning framework to jointly model the latent representations of images and text in the food image-recipe association and retrieval tasks. This model allows one to discover complex functional and hierarchical relationships between images and text, and among textual parts of a recipe including title, ingredients and cooking instructions. Our experiments show that by making use of efficient tree-structured Long Short-Term Memory as the text encoder in our computational cross-modal retrieval framework,  we are not only able to identify the main ingredients and cooking actions in the recipe descriptions without explicit supervision, but we can also learn more meaningful feature representations of food recipes, appropriate for challenging cross-modal retrieval and recipe adaption tasks.
\end{abstract}

\section{Introduction}
Computer vision and natural language processing have witnessed outstanding improvements in recent years.
Computational food analysis (CFA) 
broadly refers to methods that attempt automating food understanding, and as such, it
has recently received increased attention, in part due to its importance in health and general 
wellbeing~\cite{min_foodsurvey}. 
For instance, 
CFA  
can play an important role in assessing and learning
the functional similarity and interaction of ingredients, cooking methods and meal preferences, while aiding in computational meal preparation and planning \cite{teng2012recipe,helmy2015health}.
However, despite recent efforts CFA still poses 
specific and difficult challenges 
due to the highly heterogeneous and complex nature of the cooking transformation process.
Further to this, a particular modality may offer only a partial ``view''of the item, for example, a cooking recipe often describe elements that can easily be occluded in the visual depiction of a cooked dish, and/or come in a variety of colors, forms and textures (e.g., ingredients such as tomatoes can be green, yellow or red and can also be presented as a sauce, chunks or whole).
   
Recent approaches that aim at learning the translation between visual and textual representations of food items do so by learning the semantics of objects in a shared latent space
\cite{salvador2017,chen2018,carvalho2018,wang2019,marin2019}. Here, 
representations (also called embeddings) derived from multi-modal evidence sources (e.g., images, text, video, flavours, etc.) that belong to the same item are matched. In effect, this type of approach aims to find a common grounding language that describes items independent of their observed modality, therefore, allowing cross-modal retrieval.
Recently, recurrent neural network (RNN) architectures such as Long Short-Term Memory (LSTM) \cite{hochreiter1997} units and Gated Recurrent Units (GRU) \cite{cho2014} have (re-)emerged as two popular 
and effective models that are able to capture some long-term dependencies in sequential data.
Previous works on cross-modal image-to-recipe (test) retrieval in the food domain treat textual elements (e.g., words) as a linear sequence in a RNN \cite{salvador2017,chen2018,carvalho2018,wang2019,marin2019}.
%
However, natural language exhibits syntactic properties that would naturally combine words into phrases
in a not necessarily sequential fashion  \cite{tai2015}.
Chain structured RNNs (such as LSTMs) struggle to capture this type of relationship.
Tree-LSTM offers a generalization of LSTMs to tree-structured network topologies \cite{tai2015,zhu2015}, further to this, recent advancements in Tree-LSTMs allow online learning of the sequence structure \cite{choi2017}.
In this work, we argue that these recent advancements in Tree-LSTM structure learning are specially well suited to discover the underlying syntactic structure specific to food cooking recipes, exclusively through the signal provided by its pairing with its visual representation (food dish image).

\section{Motivation}
One of the more general goals of the work proposed here is to have a system that is able to ``understand'' food.
This is a very broad and challenging task that requires an understanding of not only the visual aspect of a meal, but also understanding what are its basic constituents and how they are processed and transformed into the final dish. Recipes offer an instruction manual as to how to prepare a dish; they specify the building blocks (ingredients), how to process them (instructions) and a succinct summary of the dish in the form of a title. Additionally, an image of a food also provides a type of dish summary in visual form, e.g., it is often possible to see main ingredients and deduce principal cooking techniques from them. As our main task is to delve deeper in the understanding of food, we would naturally focus on as many representations as possible, however, in this work we will focus primarily on recipes and images.

Some key questions that people regularly ask themselves regarding food are: what it is and how it is made.
In the proposed framework we aim at learning distinct textual entities that underline a particular dish in an unsupervised way. The proposed framework, driven solely by information arising by paired data (images and recipes), is able to understand concepts such as what is the main ingredient in this dish, thus answering the ``what it is''. This information can be 
valuable to recommendation systems, which would benefit from understanding what is the main ingredient in a recipe. For example, a person querying for apple recipes is unlikely to be interested in recipes where apples are only a minor ingredient. In order to facilitate this, it is important to also understand the key actions that are required to create a dish. Food preparation actions describe types of dishes, which can impact the likelihood of an ingredient being either major or minor. For example, assuming ``apple'' is in the list of ingredients of a dish, while showing the action ``bake'' as prominent,  it is more likely that apples are main ingredient as opposed to a recipe that where the action ``grill'' is the most prominent.
Furthermore, the ability to deeply understand food, through simple pairing of images and recipes, enables the possibility of better ``ingredient swap recipe retrieval'', where the goal is to find the dishes similar to the one described except for the main ingredient, which is replaced.

To address some of the aforementioned challenges, we propose a novel cross-modal retrieval computational framework that can effectively learn translations between images of food dishes and their corresponding preparation recipes' textual descriptions. Special emphasis is given to the functional and hierarchical relationship between text and images through the use of Tree-LSTMs. We show that using Tree-LSTMs offers not only a better representation of sentences in the context of cross-modal retrieval, but also allow us to discover important aspects of the data, that are normally lost in sequential RNNs. 

In summary our contributions are: (1) a hierarchical cross-modal retrieval framework that allows the discovery of important syntactic concepts, such as ingredient importance, keywords and/or action words, exclusively through visual-text pairings, while also providing (2) extensive experiments that demonstrate state-of-the-art performance in the image-to-recipe retrieval task as well as various recipe modifications enabled by Tree-LSTM.
Source code of our proposed method is available at \textit{https://github.com/haixpham/CHEF}.
%


\section{Cross-Modal Association Model}
\label{sec:association_model}
\begin{figure*}[!ht]
    \centering
    \includegraphics[width=0.8\textwidth]{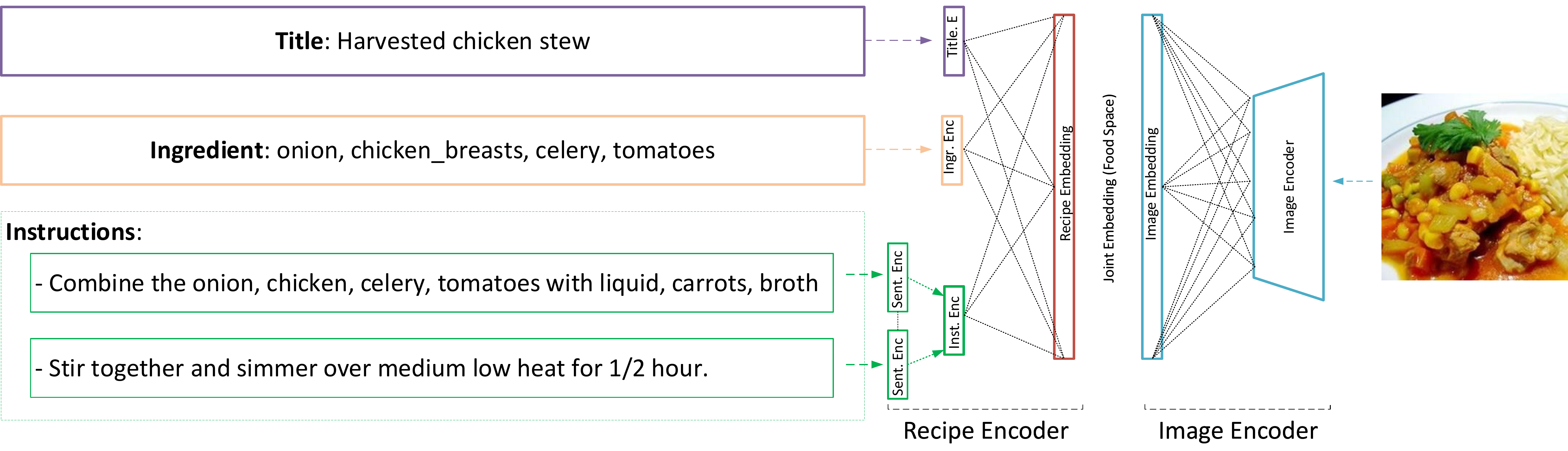}
     \caption{The general cross-modal retrieval framework, including the image encoder and recipe encoder. Within the recipe encoder, each sentence is encoded by a sub-network, and their outputs are further mapped into instruction-level latent features to be concatenated with title and ingredient embeddings.}
    \label{fig:retrieval_fw}
\vspace{-1em}
\end{figure*}

Cross-modal learning is an active research topic in computer science. In general terms, it describes a system that given a view or modality (e.g., image) of an instance, it retrieves the same instance but as viewed in another modality (e.g., text). 
These type models are usually trained using a direct correspondence between pairs of instances in different modalities. In the case of food recipe retrieval, these modalities are usually food images and their associated text descriptions (title, ingredients, recipe, etc.). 

In order to extract feature representations from both images and recipes (text), we base our architecture on a simplified version of the cross-modal association model presented by Salvador et al. \cite{salvador2017}. Different to their model, we additionally make use of titles and replace the pre-embedding of instructions with an online instruction embedding module. 
Such model is trained to match recipes (a concatenation of the encoding of title, ingredients and instructions) and their corresponding images in a joint latent space. Our general cross-modal framework is shown in Fig.~\ref{fig:retrieval_fw}.
During training, the model's objective is formulated as the minimization of the distance between an anchor recipe $\txtReal$ and matching image $\imgReal$, while also maximizing (up to a margin $\epsilon$) the distance between the anchor recipe $\txtReal$ and a non-matching image $\imgWrong$, that is, it minimizes the margin triplet loss of $(\txtReal, \imgReal, \imgWrong)$. Using two separate neural networks, one for text encoding $\TxtEnc$ and another for image encoding $\ImgEnc$, each item of the triplet is embedded in a latent space with coordinates $(\txtRealFeat, \imgRealFeat, \imgWrongFeat)$. 

Formally, with the text encoder $\txtFeat = \TxtEnc(\txt)$ and image encoder $\imgFeat = \ImgEnc(\img)$, the training is a minimization of the following objective function, 
\begin{align}
\begin{split}
    & V(\TxtEnc, \ImgEnc) = \\ 
    & \mathbb{E}_{ \hat{p}(\txtReal,\imgReal), \hat{p}(\imgWrong) } \min\left( \left[ d{\left[\txtRealFeat, \imgRealFeat\right]} - d{\left[\txtRealFeat, \imgWrongFeat\right]} - \epsilon \right], 0 \right) + \\
    & \mathbb{E}_{ \hat{p}(\txtReal,\imgReal), \hat{p}(\txtWrong) } \min\left( \left[ d{\left[\txtRealFeat, \imgRealFeat\right]} - d{\left[\txtWrongFeat, \imgRealFeat\right]} - \epsilon \right], 0 \right),
\end{split}
\label{eq:loss}
\end{align}
where $d\left[\txtFeat, \imgFeat\right] = \cos{\left[\txtFeat, \imgFeat\right]} = \txtFeat^\intercal\imgFeat / \sqrt{ (\txtFeat^\intercal\txtFeat) (\imgFeat^\intercal\imgFeat)}$ is the cosine similarity in the latent space and $\hat{p}$ denotes the corresponding empirical densities on the training set. The cosine similarity of the positive pair and that of the negative pair together are combined with a margin $\epsilon$, whose goal is to focus the model on ``hard'' examples (negatives within the margin) while ignoring those that are 
``good enough'' (beyond the margin). We empirically set $\epsilon$ to $0.3$ by cross-validation.

\subsection{LSTM Text Encoder} 
The text encoder $\TxtEnc$ takes the recipe's title, ingredients and instructions as input, and outputs their feature representation in the shared latent space. The goal is to find an embedding that reflects dependencies between a recipe's textual and visual depictions, which could facilitate implicit associations even when some components are present in only one of the two modalities, e.g., not visible but described in text or not described in text but visible. 
For this purpose, the model first converts word tokens into vectors ($w_i \in \R^{300}$) using a word2vec model \cite{mikolov2013}, treating each vector $w_i$ as part of a sequenced input to a bi-directional LSTM. The word2vec model is pre-trained on all textual information available from the recipes, then fine-tuned while training the cross-modal retrieval system.

During training, the model learns a contextual vector $\vec{h_m} \in \R^{600}$ for each of the $m$ parts of the textual information, 
that is, 
title, ingredients and 
instructions,
before concatenating them into a single 
vector $\vec{h}$.
Finally, $\vec{h}$ is projected to the shared space by three fully connected (FC) layers, each of dimensionality 1024, to yield the latent text features $\txtFeat \in \R^{1024}$. We therefore consider four LSTM modules, one each for title and ingredients, and two for instructions. The first level of the instruction encoder module encodes 
each instruction's word sequence,
then the second encodes the sequence of instructions into a single vector representation.

\subsection{Tree-LSTM Text Encoder} 
As mentioned before, the text encoder $\TxtEnc$ is tasked with encoding all textual information. 
RNNs are popular techniques to encode text,
and more specifically, LSTMs (along with GRUs) have
shown, to some degree, being able to capture the semantic meaning of text. LSTMs assume a chain graph can approximate arbitrary word dependencies within a recipe, however, this structure might not fully capture complex relationships between words. Therefore, it might be beneficial to model word dependencies as a tree structure. 
Tree-LSTM \cite{choi2017,tai2015,zhu2015} offers an elegant generalisation of LSTMs, where information flow from children to parent is controlled using a similar mechanism to a LSTM. Tree-LSTM introduces cell state in computing parent representation, which assists each cell to capture distant vertical dependencies, thus breaking the inherent linearity of a chain LSTM.
The following formulas are used to compute the model's parent representation from its children in the special case of a binary tree-LSTM:

\begin{equation}
\label{eq:tsts}
    \begin{bmatrix}
\textbf{\textup{i}}\\ 
\textbf{\textup{f}}_l\\ 
\textbf{\textup{f}}_r\\ 
\textbf{\textup{o}}\\ 
\textbf{\textup{g}}
\end{bmatrix} = 
\begin{bmatrix}
\sigma \\ 
\sigma \\ 
\sigma \\ 
\sigma \\ 
\textup{tanh}
\end{bmatrix}\left ( \textbf{\textup{W}}_{\text{comp}} 
\begin{bmatrix}
\textbf{\textup{h}}_l \\ 
\textbf{\textup{h}}_r
\end{bmatrix} + \textbf{\textup{b}}_{\text{comp}}\right )
\end{equation}

\begin{equation}
    \textbf{\textup{c}}_p = \textbf{\textup{f}}_l \odot \textbf{\textup{c}}_l + \textbf{\textup{f}}_r \odot \textbf{\textup{c}}_r + \textbf{\textup{i}} \odot \textbf{\textup{g}}
\end{equation}

\begin{equation}
    \textbf{\textup{h}}_p  = \textbf{\textup{o}} \odot \textup{tanh}\left ( \textbf{\textup{c}}_{\textbf{\textup{p}}} \right )
\end{equation}

\noindent where $\sigma$ is the sigmoid activation function,
the pairs $\left \langle \textbf{\textup{h}}_l, \textbf{\textup{c}}_l \right \rangle$ and $\left \langle \textbf{\textup{h}}_r, \textbf{\textup{c}}_r \right \rangle$
are the two input tree nodes popped off the stack,
$\textbf{\textup{W}}_{\text{comp}} \in \mathbb{R}^{5D_h \times 2D_h}$, $\textbf{\textup{b}}_{\text{comp}} \in \mathbb{R}^{2D_h}$and $\odot$ is the
element-wise product. 
The result of this function, the pair $\left \langle \textbf{\textup{h}}_\textup{\textbf{p}}, \textbf{\textup{c}}_\textup{\textbf{p}} \right \rangle$, is placed back on the stack. 
Note that the formulation used here follows that of \cite{choi2017}, which in turn is similar to \cite{bowman2016}. 

\eqref{eq:tsts} is similar to that of traditional LSTM equations except that instead of a single forget gate, there is one for each possible child of a node in the tree. More specifically to the formulation here, there are two forget gates, $\textbf{\textup{f}}_l$ and $\textbf{\textup{f}}_r$, corresponding to left and right children in a binary tree.
In this work, we make use of the Gumbel-softmax Tree-LSTM model~\cite{choi2017}, which can learn the tree structure representation without supervision.

\subsection{Image encoder} 
The image encoder $\ImgEnc$ takes an image as input and generates its feature representation in the shared latent space. ResNet50 \cite{he2016} pre-trained on ImageNet is used as the backbone for feature extraction, where the last FC layer is replaced with three consecutive FC layers (similar to the recipe encoder) to project the extracted features into the shared latent space to get $\imgFeat \in \R^{1024}$. Particularly, the middle FC layer is shared with the recipe encoder in order to preliminarily align the two modalities' distributions.

\section{Experiments}
\label{sec:exp}
In this section we will use L, T, G, and S as shorthand for LSTM, Tree-LSTM, GRU and Set~\cite{deepsets}, respectively. In this work, we use GRU to encode short sequences and LSTM for longer ones. Thus, the title encoder is chosen as G through out our experiments. The ingredient encoder is chosen amongst \{S, G, T\}, where S means all ingredients contribute equally to the final embedding. The sentence and instruction encoders are either L or T.
Different configurations of the recipe encoder are shown in Tab.~\ref{tab:xretrieval} and subsequent tables, in which a 3-tuple such as \textbf{[G+T+L]} means the model uses GRU as ingredient encoder, Tree-LSTM for sentence encoder and LSTM for the instruction-level encoder. All models are trained end-to-end.
Particularly, the attention-based model introduced by Chen et al.~\cite{chen2018} is similar to one variant of our 
framework, \textbf{[G+L+L]}, where the attention mechanism is integrated into every text encoder. However, this model is trained within our experimental setting, which lead to improved performance in general. 

We evaluate our proposed models in four different tasks, including (i) cross-modal retrieval, (ii) main ingredients detection (including ingredient pruning), (iii) ingredient substitution and (iv) action words extraction. Additionally, we compare the performance of our models to that of the original R1M pic2rec model~\cite{salvador2017} and the state-of-the-art  ACME~\cite{wang2019}, retrained to adapt to our vocabulary (cf. Sec.~\ref{sec:exp:dataset}), while faithfully following their original settings.
We also retrained Adamine~\cite{carvalho2018} using publicly available source code, however, we were unable to make it work adequately with our modified data, thus we did not include it in this paper.

\subsection{Dataset}
\label{sec:exp:dataset}
During the preparation of the work presented here, all experiments were conducted using data from Recipe1M (R1M) \cite{salvador2017,marin2019}. 
This dataset consists of $\sim$1M text recipes that contain titles, instructions and ingredients. Additionally, a subset of $\sim$0.5M recipes contain at least one image per recipe. 
Data is split into 70\% train, 15\% validation and 15\% test sets. During training at most 5 images from each recipe are used, while a single image is kept for validation and testing.

\noindent{\bf Canonical Ingredient Construction.}
To analyze ingredient relative importance across different recipes, a standardized version of R1M ingredients was created.
R1M contains $\sim$16k unique ingredients, with the top 4k accounting for $\sim$95\% occurrences. Focusing on these, we 
reduced them to $\sim$1.4k through the following operations.
%
First, ingredients are merged if they have the same name after 
stemming and singular/plural conversion.
Second, ingredients are merged if they are close together in our word2vec \cite{mikolov2013} embedding space,
if they share two or three words or are mapped to the same item by Nutritionix\footnote{https://www.nutritionix.com/}. Lastly, 
mergers are 
inspected by a human 
who can 
accept or reject 
them.

\subsection{Cross-modal Recipe Retrieval}
\label{sec:exp:xretrieval}

 \begin{table}[!ht]
  \centering
  \scriptsize
  \captionsetup{font=small}
  \caption{Image-to-Recipe retrieval performance comparison between model variants of our proposed framework and the baselines. The Recipe-to-Image retrieval performance is similar, and is included in the appendix. The models are evaluated on medR (lower is better) and Recall@K (R@K - higher is better). In this table and subsequent tables, our proposed models are ordered by type of ingredient - sentence - instruction encoders. Best results are marked in bold.
  }
     \begin{tabular}{lcccc}
     \toprule
     \textbf{Methods} & \textbf{medR}$\downarrow$ & \textbf{R@1}$\uparrow$ & \textbf{R@5}$\uparrow$ & \textbf{R@10}$\uparrow$ \\ %
     \midrule
          \multicolumn{5}{c}{\textbf{Size of test set: 1k}} \\
     \midrule
    pic2rec~\cite{salvador2017}   & 4.10  & 26.8  & 55.8  & 67.5  \\
    ACME~\cite{wang2019}  & 2.00  & 45.4  & 75.0  & 83.7  \\
    S+L+L & 1.80  & 48.0  & 77.0  & 84.1  \\
    S+T+T & 2.20  & 37.7  & 68.0  & 78.7  \\
    G+L+L~\cite{chen2018} & \textbf{1.60} & 49.3  & 78.1  & 85.2  \\
    G+T+L & 1.80  & 49.0  & 78.0  & 85.8  \\
    G+T+T & 1.80  & 48.7  & 78.3  & 85.7  \\
    T+L+L & 1.80  & 49.4  & \textbf{79.6} & 86.1  \\
    T+L+T & \textbf{1.60} & \textbf{49.7} & 79.3  & \textbf{86.3} \\
    T+T+L & 1.75  & 49.0  & 78.8  & 85.9  \\
    T+T+T & 1.70  & 49.4  & 78.8  & 85.9  \\
    \midrule
          \multicolumn{5}{c}{\textbf{Size of test set: 10k}} \\
    \midrule
    pic2rec~\cite{salvador2017}   & 33.25 & 7.7   & 21.8  & 30.8  \\
    ACME~\cite{wang2019}  & 9.40  & 18.0  & 40.3  & 52.0  \\
    S+L+L & 8.10  & 19.6  & 42.8  & 54.5  \\
    S+T+T & 15.90 & 13.2  & 31.8  & 42.8  \\
    G+L+L~\cite{chen2018} & 7.50  & 20.7  & 44.7  & 56.2  \\
    G+T+L & 7.60  & 20.7  & 44.3  & 55.9  \\
    G+T+T & 7.50  & \textbf{20.9} & 44.5  & 56.0  \\
    T+L+L & \textbf{7.30} & \textbf{20.9} & \textbf{44.8} & \textbf{56.3} \\
    T+L+T & \textbf{7.30} & 20.7  & 44.7  & 56.2   \\
    T+T+L & 7.50  & 20.8  & 44.3  & \textbf{56.3}  \\
    T+T+T & \textbf{7.30} & \textbf{20.9} & 44.6  & 56.1  \\
    \bottomrule
     \end{tabular}%
    
  \label{tab:xretrieval}%
 \vspace{-1em}
 \end{table}

\begin{table}[!ht]
  \centering
  \scriptsize
  \captionsetup{font=small}   
  \caption{Main ingredient prediction performance. The models evaluated on medR (lower is better) and Recall@K (R@K - higher is better).}
     \begin{tabular}{llcccc}
     \toprule
     \textbf{Model}&\textbf{Split} & \textbf{medR}$\downarrow$ &  \textbf{R@1}$\uparrow$ & \textbf{R@2}$\uparrow$ & \textbf{R@3}$\uparrow$ \\ %
     \midrule
     \multirow{2}{*}{Tree-LSTM} & Test      & 1.0  &  47.0  & 78.9  & 95.5 \\
                                & Validation& 1.0  &  47.3  & 79.7  & 95.5 \\ 
     \cmidrule(lr){1-1} \cmidrule(lr){2-2} \cmidrule(lr){3-6}
     \multirow{2}{*}{Attention} & Test      & 4.0  &  9.1   & 19.5  & 32.1 \\
                                & Validation& 4.0  &  9.5   & 19.7  & 32.2 \\
     \bottomrule
     \end{tabular}%
  \label{tab:main_ingr}%
 \vspace{-1em}
 \end{table}%

The cross-modal retrieval performance of our proposed models are compared against the baselines in Tab.~\ref{tab:xretrieval}. We report results on two subsets of randomly selected 1,000 and 10,000 recipe-image pairs from the test set, similar to \cite{wang2019}. These experiments are repeated 10 times, and we report the averaged results. All variants of our proposed framework, except \textbf{[S+T+T]}, outperform the current state-of-the-art, ACME - retrained on the modified dataset, across all evaluation metrics. Interestingly, the attention-based model~\cite{chen2018}, when trained within our paradigm, achieves significant gain over ACME, and it scores the best median rank (\textit{medR}) on the recipe-to-image retrieval task. It is worth noting that, we were unable to reproduce the results of ACME as reported by the authors on the original, unmodified dataset (more analysis of ACME performance included in the appendix).

Our proposed models have similar performance scores, and amongst those, \textbf{[T+L+T]} is the best performer in most cases. More importantly, these empirical evidences where \textbf{[T+L+L]} and \textbf{[G+L+L]} perform better than \textbf{[S+L+L]} suggest that ingredients hold significant information to recognize a dish from its image, thus a more effective ingredient encoder like Tree-LSTM or GRU with attention will be able to embed more meaningful information in the food space, which also improves the capability of the image encoder when they are trained jointly using our proposed triplet loss~\eqref{eq:loss}, which is not regularized by semantics information, unlike pic2rec and ACME. Furthermore, Tree-LSTM explicitly imposes implicit importance scores to different ingredients due to its inherent hierarchical structure, hence this observation encourages us to analyze the roles of ingredients in the joint food space more thoroughly in the next two sections. Sec.~\ref{sec:exp:main_ingr_disc} will also show that Tree-LSTM is better at attending to the important ingredients than the soft attention mechanism.

\subsection{Main Ingredients Detection}
\label{sec:exp:main_ingr_disc}

One benefit of using Tree-LSTM to encode text is that, the information flow in the network follows a learned hierarchical structure, and the closer the word (its corresponding leaf node in the tree) is to the root, the more importance it holds in the text. Thus, intuitively when embedding the list of ingredients, Tree-LSTM should be able to infer a hierarchy that emphasizes the more important ingredients, including the main one. We extract tree structures of all ingredient embeddings generated by Tree-LSTM in the test set, and calculate their average depths. 

\begin{figure}[!ht]
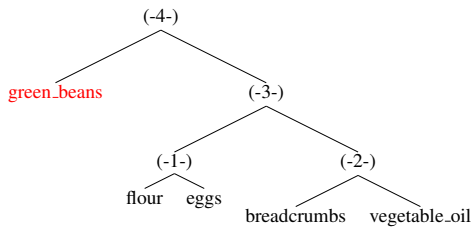

	\centering
	
	\resizebox{0.35\textwidth}{!}{
		\Tree [.(-4-)
		\textcolor{red}{green\_beans}
		[.(-3-)
		[.(-1-) flour eggs ]
		[.(-2-) breadcrumbs vegetable\_oil ] ] ]
	}
	\caption{Ingredient tree of ``Deep fried green beans''. The order of flattened leaves from left to right is the same order they appear in the original ingredient list. The labels of intermediate nodes indicate the order of word combinations by Tree-LSTM, the higher value means its children are combined later. In this example, ``flour'' and ``eggs'' are combined first, and ``green\_beans'' is embedded last into the final ingredient embedding.}
	\label{fig:main_ingr}
	\vspace{-1em}
\end{figure}

We observed that all top ingredients have the average depths in the range of 2.x, which is very close to the root.
In other words, the Tree-LSTM ingredient encoder has learned the hierarchical representations that impose higher importance on some ingredients. An example of ingredient tree is illustrated in Fig.~\ref{fig:main_ingr}, where the main ingredient, ``green\_breans'' is at the lowest leaf of the tree, meaning it is the most important ingredient of this dish, as the model has learned. More examples are included in the supplementary materials.
However, these observations still do not answer the question, are these top ingredients always the main ones in their respective recipes?

We conjecture that the main ingredient of each recipe is likely the one with the largest mass among all ingredients. In order to validate this intuition, a subset of $\sim$30k test and validation samples (15k each) from Recipe1M has been curated to include ingredient amounts and units.
In this data, ingredient amounts and units have been normalized to a standard total weight of 1kg, e.g. fluid units were converted to weight. An experiment was designed where the gold standard for a recipe's main ingredient was assumed to be well aligned with the amount by weight. Therefore, such ingredient should be predicted as the shallowest during Tree-LSTM inference. From the Tree-LSTM embedding of each of these recipes we can quantify the rank of the \textit{true} main ingredient with regards to the root, that is, what is the distance between the true main ingredient and the shallowest ingredient in the tree. Tab.~\ref{tab:main_ingr} summarizes these results, and as it can be seen in $\sim$47\% of the cases the correct main ingredient is identified. 
Given the median recipe contains 9 ingredients (with maximum ingredient depth of 8), a random na\"ive predictor would rank ingredients as 4, $\sim$95\% of our predicted ranks are better than chance. This table also includes the prediction performance of the attention-based approach in\cite{chen2018}, which one would expect to ``attend'' to the main ingredient in a recipe. However, as the results indicate, the attention mechanism fails to uncover the main ingredients.

\begin{table}[!ht]
\centering
\scriptsize
\captionsetup{font=small}
\caption{Retrieval performance of \textbf{[T+T+L]} after pruning K ingredients corresponding to K highest leaves in the ingredient tree. The results are averaged over 10 runs of 1k random recipes each.}
\label{tab:prune_perf}
\begin{tabular}{ccccccc}
\toprule
\multicolumn{1}{l}{} & \multicolumn{3}{c}{\textbf{Image-to-Recipe}}      & \multicolumn{3}{c}{\textbf{Recipe-to-Image}}      \\
\textbf{K}           & \textbf{medR}$\downarrow$ & \textbf{R@1}$\uparrow$ & \textbf{R@5}$\uparrow$  & \textbf{medR}$\downarrow$ & \textbf{R@1}$\uparrow$ & \textbf{R@5}$\uparrow$  \\
\midrule
0                    & 1.75          & 49.0          & 78.8          & 1.60          & 49.7          & 78.9   \\
1                    & \textbf{1.10} & \textbf{51.1} & \textbf{79.9} & \textbf{1.00} & \textbf{51.9} & \textbf{79.9} \\
2                    & 1.50          & 50.0           & 79.4          & 1.20          & 50.8          & 79.3  \\
3                    & 1.85          & 48.7          & 78.2           & 1.80          & 49.4          & 77.8   \\
4                    & 1.90          & 46.3          & 76.6          & 1.95          & 47.8          & 76.6  \\
\bottomrule
\end{tabular}
\end{table}

\subsubsection{Ingredient Pruning}
As Tree-LSTM ingredient encoder has the unique ability to address the importance of ingredients, a set of experiments were conducted in which we removed the K least important ingredients in the list, corresponding to the K highest leaves in the tree, and their occurrences in the instructions were also removed. The retrieval performance of these novel (ingredient pruned) recipes using the \textbf{[T+T+L]} recipe encoder is demonstrated in Tab.~\ref{tab:prune_perf}. This table shows that removing the least important ingredient of each recipe (K=1) actually brought significant performance gain: medR of image-to-recipe and recipe-to-image retrievals are 1.1 and 1.0, respectively, while R@1 improves by 2 points. Removing two ingredients also improves retrieval performance, without learning a new model. The performance after removing three ingredients is roughly equal to that of the original recipes, and performance starts decreasing after removing four ingredients, which makes sense as on average four ingredients constitute 25\% of the ingredient list. Overall, these results reaffirm the ability of Tree-LSTM ingredient encoder to attend to the more important ingredients in the recipes, and suggest the exploration of a self-pruning recipe encoder that may improve performance of downstream tasks.

\begin{table*}[!ht]
  \centering
  \scriptsize
  \captionsetup{font=small}
  \caption{Substitution from ``chicken'' to other ingredients. The values shown are Success Rate (SR) (higher is better). R-2-I and R-2-R indicate novel recipe-to-image and novel recipe-to-recipe retrievals, respectively.
  The last column shows the median ranks of all models in terms of SR across all experiments (lower is better). Best results are marked in bold.}
    \begin{tabular}{lccccccccc}
    \toprule
           & \multicolumn{2}{c}{\textbf{To Beef}} & \multicolumn{2}{c}{\textbf{To Apple}} & \multicolumn{2}{c}{\textbf{To Pork}} & \multicolumn{2}{c}{\textbf{To Fish}} &  \\
    \cmidrule(lr){1-1} \cmidrule(lr){2-3} \cmidrule(lr){4-5} \cmidrule(lr){6-7} \cmidrule(lr){8-9} \cmidrule(lr){10-10}
    \textbf{Methods} & \textbf{R-2-I} & \textbf{R-2-R} & \textbf{R-2-I} & \textbf{R-2-R} & \textbf{R-2-I} & \textbf{R-2-R} & \textbf{R-2-I} & \textbf{R-2-R} & \textbf{Med. Rank} \\
    \cmidrule(lr){1-1} \cmidrule(lr){2-3} \cmidrule(lr){4-5} \cmidrule(lr){6-7} \cmidrule(lr){8-9} \cmidrule(lr){10-10}
    pic2rec~\cite{salvador2017}    & 17.3   & \textbf{22.5}   & 5.9    & 9.0    & 10.6   & \textbf{18.9} & 2.7    & 2.9    & 10.0 \\
    ACME~\cite{wang2019}   & 18.6   & 18.9 & 5.4    & 6.3    & 8.2   & 8.9   & 3.3    & 3.1    & 10.0 \\
    S+L+L & 29.6   & 16.1   & 10.3   & 8.4    & \textbf{15.9} & 6.3    & \textbf{8.6} & 3.5    & 5.5 \\
    S+T+T & 29.9   & 7.3    & 9.3    & 7.2    & 13.9   & 4.1    & 5.9    & 2.6    & 9.5 \\
    G+L+L \cite{chen2018}  & 28.5   & 16.2   & 9.2    & 7.8    & 15.2   & 5.8    & 7.1    & 3.1    & 8.0 \\
    G+T+L & 29.2   & 16.2   & 9.9    & 9.2    & 15.3   & 7.1    & 7.5    & 3.6    & 6.0 \\
    G+T+T & 28.2   & 17.6   & 11.4   & 9.4    & 13.9   & 7.1    & 6.9    & 3.1    & 6.5 \\
    T+L+L & 29.8   & 20.0   & 12.0   & 10.4   & 14.5   & 7.0    & 6.0    & 3.5    & 5.0 \\
    T+L+T & 29.6   & 20.9   & 11.3   & 11.3   & 15.0   & 7.4    & 7.4    & 4.1    & 4.0 \\
    T+T+L & \textbf{31.0} & 21.1   & 11.7   & 9.7    & 13.8   & 7.6    & 7.0    & \textbf{4.2} & 3.5 \\
    T+T+T & 30.0   & 20.3   & \textbf{12.2} & \textbf{11.7} & 15.4   & 7.8    & 6.5    & 4.0    & \textbf{2.5} \\
    \bottomrule
    \end{tabular}%
  \label{tab:Subs_from_chicken}%
\end{table*}%

\begin{table*}[!ht]
  \centering
  \scriptsize
  \captionsetup{font=small}
  \caption{Substitution from ``pork'' to other ingredients. Best results are marked in bold.}
    \begin{tabular}{lccccccccc}
    \toprule
           & \multicolumn{2}{c}{\textbf{To Chicken}} & \multicolumn{2}{c}{\textbf{To Beef}} & \multicolumn{2}{c}{\textbf{To Apple}} & \multicolumn{2}{c}{\textbf{To Fish}} &  \\
    \cmidrule(lr){1-1} \cmidrule(lr){2-3} \cmidrule(lr){4-5} \cmidrule(lr){6-7} \cmidrule(lr){8-9} \cmidrule(lr){10-10}
    \textbf{Methods} & \textbf{R-2-I} & \textbf{R-2-R} & \textbf{R-2-I} & \textbf{R-2-R} & \textbf{R-2-I} & \textbf{R-2-R} & \textbf{R-2-I} & \textbf{R-2-R} & \textbf{Med. Rank} \\
    \cmidrule(lr){1-1} \cmidrule(lr){2-3} \cmidrule(lr){4-5} \cmidrule(lr){6-7} \cmidrule(lr){8-9} \cmidrule(lr){10-10}
    pic2rec~\cite{salvador2017}    & 41.2   & \textbf{47.4} & 25.6   & 27.6   & 6.8    & 13.7   & 3.2    & 4.2    & 11.0 \\
    ACME~\cite{wang2019}   & 30.0   & 29.8   & 30.7   & \textbf{28.4} & 7.1   & 9.7   & 2.5    & 3.7    & 10.0 \\
    S+L+L & 47.2   & 23.0   & 39.6   & 17.7   & 19.6   & 21.3   & \textbf{9.0} & 6.2    & 5.0 \\
    S+T+T & 47.7   & 17.0   & 41.0   & 12.8   & 13.4   & 18.0   & 7.4    & 4.4    & 9.0 \\
    G+L+L \cite{chen2018} & 47.9   & 23.6   & 40.8   & 17.2   & 14.6   & 17.5   & 7.0    & 5.4    & 5.0 \\
    G+T+L & 47.8   & 24.6   & 39.5   & 18.6   & 14.4   & 20.6   & 8.5    & 6.4    & 5.0 \\
    G+T+T & 46.4   & 24.8   & 39.5   & 21.1   & 17.5   & 20.3   & 7.8    & 5.5    & 5.0 \\
    T+L+L & \textbf{49.7} & 29.9   & 42.9   & 21.1   & 18.1   & 20.3   & 6.9    & 5.3    & \textbf{3.0} \\
    T+L+T & 44.8   & 29.2   & 40.9   & 22.7   & 17.8   & 22.8   & 8.6    & 6.5    & 4.0 \\
    T+T+L & 47.6   & 28.4   & 42.0   & 22.1   & \textbf{20.3} & 19.4   & 8.4    & \textbf{7.5} & 6.0 \\
    T+T+T & 45.7   & 27.7   & \textbf{43.9} & 23.0   & 16.9   & \textbf{23.1} & 6.0    & 6.2    & 4.0 \\
    \bottomrule
    \end{tabular}%
  \label{tab:subs_from_pork}%
\end{table*}%

\subsection{Ingredient Substitution}
\label{sec:exp:ingr_subs}

\begin{figure*}[!ht]
\centering
    \subfloat[Modifying grilled chicken to grilled beef]{
        \includegraphics[width=0.45\textwidth]{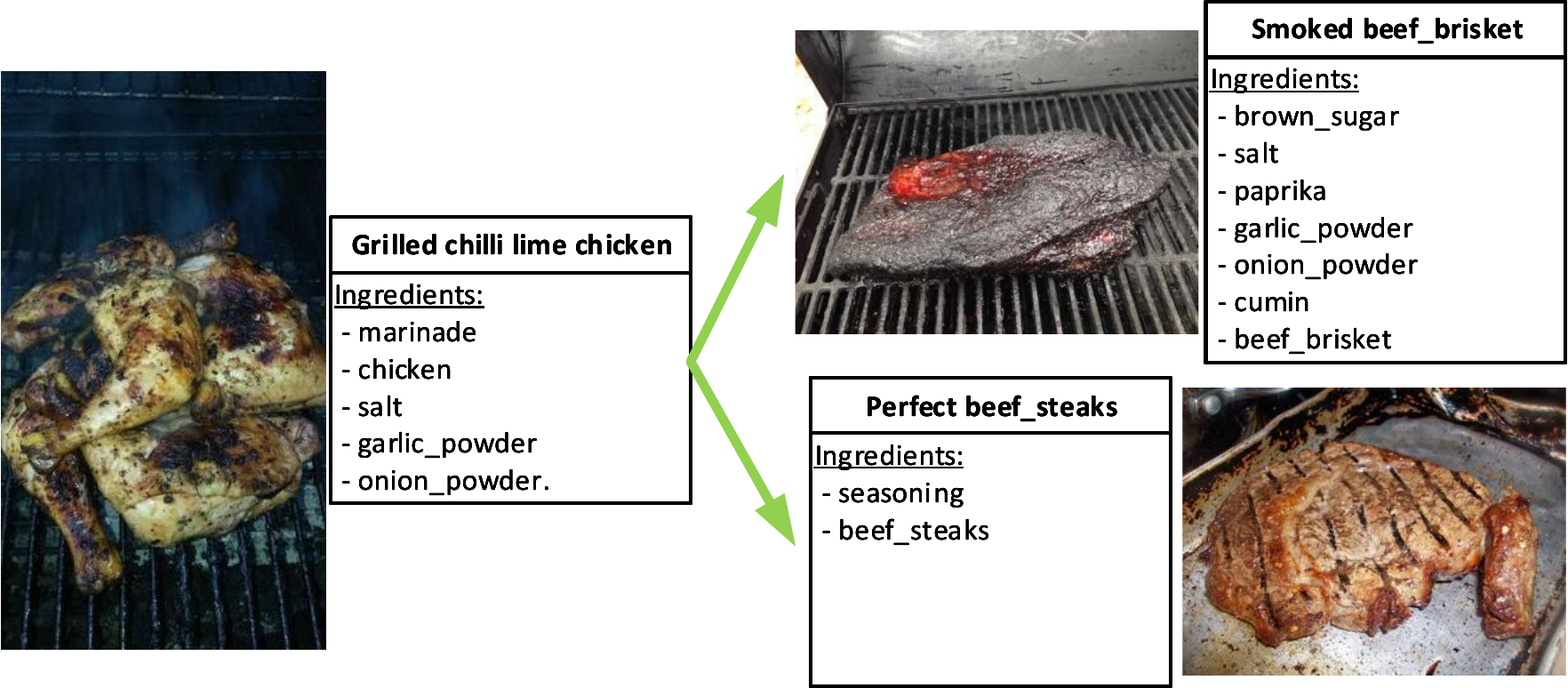}
        \label{fig:grilled_chicken2beef}
    }
	\hspace{2em}
    \subfloat[Modifying grilled chicken to fried apple]{
        \includegraphics[width=0.45\textwidth]{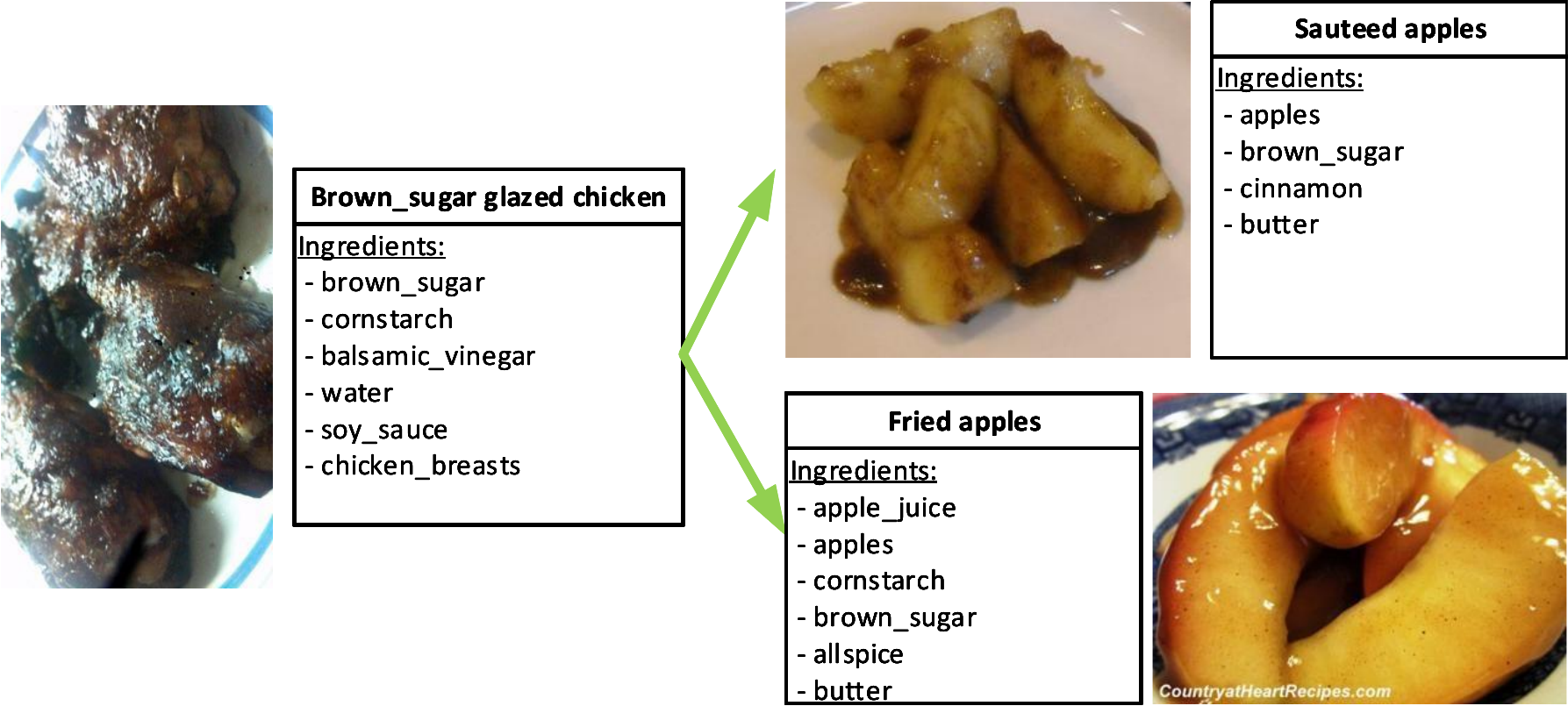}
        \label{fig:chicken2apple}
    }
\caption{Ingredient substitution and recipe retrieval. ``Chicken'' in each recipe is replaced with ``beef'' or ``apple'' and carry out retrievals on both image and text embeddings. In each sub-figure: top-right and bottom-right show top-1 image retrieval and top-1 recipe text retrieval, respectively.}
\label{fig:ingr_subs}
\vspace{-1.0em}
\end{figure*}

The above observation that Tree-LSTM can put more importance upon main ingredients leads to one interesting exploration: if the main ingredient of a recipe were replaced with another, would the embedding of the new recipe reflects this change, e.g., by retrieving real recipes containing this new ingredient? Inspired by earlier research on this task~\cite{Shidochi2009,Tsukuda2010,Yokoi2015,Cordier2013}, we carry out the substitutions by directly replacing the ingredient token in the input recipe with another. However, unlike prior endeavors which rely on standard text ranking, in this work we utilize deep models to extract the embeddings that can be used to retrieve \textit{both} images and recipes. We propose a new metric to compare substitution performance of different models, namely \textit{success rate} (SR), defined as the percentage of recipes containing ingredient $A$ that were successfully identified as containing ingredient $B$ by retrieval on a dataset, after replacing $A$ with $B$ in these recipes. Moreover, we report results where the original tree structures of the recipes are retained when inferring the new embeddings. There are marginal changes in performance when we let the text encoders infer new hierarchies from the modified recipes, suggesting that the new structures are similar to the originals in most cases.

We select recipes containing ``chicken'' and ``pork'' where they are identified as main ingredient by the Tree-LSTM ingredient encoder, and replacing them with one of the following ingredients where applicable: ``chicken'', ``beef'', ``pork'', ``fish'' and ``apple''. These ingredients are recognized as top ingredients by Tree-LSTM encoder. The results are shown in Tab.~\ref{tab:Subs_from_chicken} and~\ref{tab:subs_from_pork}.

In the experiments converting chicken-based recipes to beef- and pork-based dishes, we see two contradicting results between image retrievals and text retrievals. In the R-2-I task, our proposed models outperform pic2rec and ACME by $\sim$50\%, however, in the R-2-R task, pic2rec and ACME perform better, especially in the case of ``pork'' conversion. We observe that in the case of ``pork'' recipes, the titles usually do not contain the word ``pork'', but title contributes 33\% of information to the final recipe embedding. Thus, the novel embeddings encoded by our models do not move to the correct manifold of ``pork'' recipes. On the other hand, pic2rec and ACME models do not include titles in the embeddings, hence they can match the novel recipes with real recipes better. It also suggests that the image embeddings are influenced more by ingredients in the corresponding recipes and their titles, hence our models perform better in the R-2-I task. It is also commonly observed that the meal images often expose the main ingredients. This explains why our proposed models perform better in more unrealistic substitutions to ``fish'' and particularly ``apple'', as these substitutions may generate nonsensical recipes, however, these ingredients are more effectively emphasized by using the tree hierarchies, thus the final embeddings are moved towards the respective manifolds of the substituting ingredients. 
The overall median ranks indicate that our proposed models, which use TreeLSTM ingredient encoder, are generally the better performers. Similar conclusions can be deduced from Tab.~\ref{tab:subs_from_pork}. These results suggest that \textbf{[T+T+L]} and \textbf{[T+T+T]} perform consistently well across different ingredient substitution experiments. These results also show that using Tree-LSTM as sentence and instruction encoders does not really have an effect on boosting successful substitution rates.

Fig.~\ref{fig:ingr_subs} demonstrates replacing ``chicken'' with ``beef'' and ``apple''.  Replacing ``chicken'' with ``beef'' in the original recipe, ``grilled chicken'', in Fig.~\ref{fig:grilled_chicken2beef}, will match with real ``grilled beef'' recipes. In Fig.~\ref{fig:chicken2apple}, replacing ``chicken'' will retrieve ``fried apple''. This suggests that the cooking methods of the original recipes are preserved in the modified embeddings. In the next section, we will investigate whether the Tree-LSTM sentence encoder can capture the cooking methods, i.e., emphasize the importance of action words.

\subsection{Action Word Extraction}
\label{sec:exp:action_word}

\begin{table}[!ht]
\centering
\scriptsize
\captionsetup{font=small}
\caption{Number of action words as the lowest leaves and their percentage over the number of sentence trees.}
\label{tab:actionword}
\begin{tabular}{ccccc}
\toprule
{} & \multicolumn{2}{c}{Validation Set} & \multicolumn{2}{c}{Test Set} \\ \hline
\multicolumn{1}{c}{Models} & Verb Count & \% & Verb Count & \% \\ \midrule
G+T+L & \textbf{229,042} & \textbf{78.44} & \textbf{227,840} & \textbf{78.20} \\ 
T+T+L & 224,709 & 76.96 & 224,550 & 77.07 \\ 
S+T+T & 185,877 & 63.66 & 185,599 & 63.70 \\ 
G+T+T & 164,436 & 56.32 & 163,432 & 56.09 \\ 
T+T+T & 139,972 & 47.94 & 139,499 & 47.88 \\ \bottomrule
\end{tabular}
\vspace{-1em}
\end{table}

Previous sections have demonstrated that it is possible to discover main ingredients with unsupervised training in our cross-modal retrieval framework. However, a cooking recipe not only consists of ingredients, but also describes a series of actions applied to ingredients in order to prepare the meal. Thus, it is equally important to emphasize these key actions to understand the recipe. An RNN model encodes a sentence sequentially, hence it is unable to specifically focus on a word arbitrarily positioned in the sentence. This problem can be partially remedied by applying the attention mechanism~\cite{Bahdanau2015,vaswani2017}, however, Sec.~\ref{sec:exp:main_ingr_disc} demonstrates lack of correlation with importance of words. Tree-LSTM, on the other hand, provides a natural representation of sentence structure.

We investigate different recipe encoders trained with our cross-modal retrieval objective, in which sentences are encoded using Tree-LSTM, while the ingredient and instruction encoder sub-networks vary. Tree structures of all sentences in the validation and test sets are thoroughly analyzed. Based on the intuition that the leaf closest to the root of the tree might be the most important word, and that a sentence in a cooking recipe typically has one action word - a verb, we collect all leaf nodes closest to the root across all sentence trees, and count how many of them are verbs appearing in the WordNet database. 

The results in Tab.~\ref{tab:actionword} demonstrate that two models using Tree-LSTM to encode sentences and LSTM to encode the whole instructions are able to emphasize on the main action words in more than 76\% of the number of sentences.
\textbf{[G+T+L]} is marginally better than \textbf{[T+T+L]}. 
It can be explained that when Tree-LSTM is used for ingredient encoder, the recipe encoding network learns to focus more on ingredients, thus the importance of instructions is somewhat subsided. It is also noticeable that performance of models using Tree-LSTM to encode the whole instruction significantly declines. This is perhaps because sentences in a recipe are usually written in chronological order, hence learning instruction-level Tree-LSTM actually is detrimental to the ability to encode action words. 
Examples of inferred sentence trees are included in the appendix.

\section{Conclusion}

In this paper, we present a novel cross-modal recipe retrieval framework that learns to jointly align the latent representations of images and texts. We particularly provide in-depth analysis of different architectures of the recipe encoder through a series of experiments. By using Tree-LSTM to model the hierarchical relationships between ingredients, and between words in instructional sentences, it is possible to capture more meaningful semantics from the recipe descriptions, such as main ingredient, cooking actions, thus also gaining better recipe adaptation capability and improving cross-modal retrieval performance
as demonstrated by variants of our proposed framework, especially the \textbf{[T+T+L]} model which performs consistently well across different experimental tasks.
In the future, we would like to jointly model the relationships between entities of the  visual and textual modalities.
{\small
\bibliography{egbib}
}

\clearpage
\begin{appendices}
\appendixpage

\section{Model and Training Details}
\label{sec:model_train}

\paragraph{\textbf{Words.}}
An initial 300-dimensional {\em word2vec} model~\cite{mikolov2013} is trained using all training recipes. Specifically, in this work we consider the list of pre-extracted ingredient names provided by Recipe1M~\cite{salvador2017}. This initial {\em word2vec} model is used as part of the canonical ingredient list creation. A second 300-dimensional {\em word2vec} model is trained, where all occurrences of ingredients names are replaced by their canonical form. This second model is used to initialize word embeddings, which are further updated during training. Note that words that originally appear less than 10 times are excluded from our vocabulary and replaced with the ``$<$UNK$>$'' token:

\paragraph{\textbf{Title.}}
%
%
A title is considered to be a sequence of words; it is passed in the order the words appear in the title to a bi-directional GRU~\cite{cho2014}. The dimensionality of the hidden state is 300 and both forward and backward states of the bi-directinal GRU are concatenated into a single 600-dimensional ingredients vector.

\paragraph{\textbf{Ingredients.}}
We consider three possible ways to find the ingredient list/set latent space, mainly, Dense (Set), LSTM~\cite{hochreiter1997} and Tree-LSTM~\cite{choi2017}.

\begin{itemize}
\item \textbf{Dense}: As implied, ingredients are treated as ``set''~\cite{deepsets} and passed through a fully connected layer that increases the dimensionality of each ingredient from 300 to 600. In order to combine all ingredients into a single latent state, we take the sum of all ingredient vectors as the final embedding.

\item \textbf{LSTM}: Similar to title, ingredients are seen as a sequential ``list'' and are passed in the order they appear to a bi-directional LSTM with 300-dimensional hidden states. Both directional hidden states are concatenated into a single 600-dimensional ingredients vector.

\item \textbf{Tree-LSTM}: Ingredients are seen as a ``list'' of words, however, the order is not implied but inferred during training. The dimensionality of the hidden state of this layer is set to 600.
\end{itemize}

\paragraph{\textbf{Instructions.}}
Generally, instructions represent the main body of text of the recipe, frequently are composed of hundreds of words. Therefore, a single LSTM or Tree-LSTM is not sufficient to capture their semantics. In this work, instructions are encoded first as a sequence of words per one instruction, then as a sequence of instructions per recipe. Both of these encoding steps can be again modeled as an LSTM or a Tree-LSTM, with the same dimensionality as that of the title or ingredients. 

\paragraph{\textbf{Complete Text Features.}}
Prior to finding a complete recipe embedding, the features from title, ingredients and instructions, each of dimensionality 600, are concatenated. This 1800-dimensional feature vector is passed through a fully connected (FC) layer that projects it to 1024 dimensions. Next, the embedding is passed though another FC layer shared with the image encoder, and the final FC layer to project it into the 1024-dimensional latent space. The role of the shared layer is to gradually align both image and text embeddings by aligning the mean and standard deviation of their distributions, before the final instance wise alignment in the target latent space.

\paragraph{\textbf{Model Training.}}
All models, including our version of the model proposed by Chen et al.~\cite{chen2018}, are implemented in PyTorch, and trained end-to-end using the Adam optimizer~\cite{adam} under the same settings. In particular, the word embedding is initialized with the {\em word2vec} model described above, and ResNet50 is initialized with parameters pre-trained on ImageNet. The two baseline models, pic2rec~\cite{salvador2017} and ACME~\cite{wang2019} are retrained on our vocabulary, using their published source code and training setups.

\section{Experiments}

\subsection{Cross-modal Retrieval}
\label{sec:xret_sup}

\begin{table*}[!ht]
  \centering
  \small
  \caption{Cross-modal retrieval performance comparison between model variants of our proposed framework and the baselines. The models are evaluated on medR (lower is better) and Recall@K (R@K - higher is better). In this table and subsequent tables, our proposed models are ordered by type of ingredient - sentence - instruction encoders. Best results are marked in bold.}
     \begin{tabular}{lcccccccc}
     \toprule
          & \multicolumn{4}{c}{\textbf{Image-to-Recipe}} & \multicolumn{4}{c}{\textbf{Recipe-to-Image}} \\ \midrule
     \textbf{Methods} & \textbf{medR}$\downarrow$ & \textbf{R@1}$\uparrow$ & \textbf{R@5}$\uparrow$ & \textbf{R@10}$\uparrow$ & \textbf{medR}$\downarrow$ & \textbf{R@1}$\uparrow$ & \textbf{R@5}$\uparrow$ & \textbf{R@10}$\uparrow$ \\ %
     \midrule
          & \multicolumn{8}{c}{\textbf{Size of test set: 1k}} \\
     \midrule
    pic2rec~\cite{salvador2017}   & 4.10  & 26.8  & 55.8  & 67.5  & 3.80  & 30.2  & 57.7  & 68.7 \\
    ACME~\cite{wang2019}  & 2.00  & 45.4  & 75.0  & 83.7  & 2.00  & 47.1  & 76.0  & 84.5 \\
    S+L+L & 1.80  & 48.0  & 77.0  & 84.1  & 1.70  & 48.8  & 76.8  & 84.7 \\
    S+T+T & 2.20  & 37.7  & 68.0  & 78.7  & 2.10  & 38.7  & 68.9  & 79.2 \\
    G+L+L~\cite{chen2018} & \textbf{1.60} & 49.3  & 78.1  & 85.2  & \textbf{1.30} & \textbf{50.4}  & 78.1  & 85.2 \\
    G+T+L & 1.80  & 49.0  & 78.0  & 85.8  & 1.50  & 49.9  & 78.7  & 85.9 \\
    G+T+T & 1.80  & 48.7  & 78.3  & 85.7  & 1.60  & 49.8  & 78.6  & 85.8 \\
    T+L+L & 1.80  & 49.4  & \textbf{79.6} & 86.1  & 1.45  & 49.8  & \textbf{79.0} & 86.4 \\
    T+L+T & \textbf{1.60} & \textbf{49.7} & 79.3  & \textbf{86.3} & 1.60  & 50.1 & \textbf{79.0} & 86.4 \\
    T+T+L & 1.75  & 49.0  & 78.8  & 85.9  & 1.60  & 49.7  & 78.9  & \textbf{86.6} \\
    T+T+T & 1.70  & 49.4  & 78.8  & 85.9  & 1.50  & 49.9  & 78.8  & 86.2 \\
    \midrule
          & \multicolumn{8}{c}{\textbf{Size of test set: 10k}} \\
    \midrule
    pic2rec~\cite{salvador2017}   & 33.25 & 7.7   & 21.8  & 30.8  & 28.50 & 9.9   & 25.0  & 34.2 \\
    ACME~\cite{wang2019}  & 9.40  & 18.0  & 40.3  & 52.0  & 8.75  & 19.4  & 42.2  & 53.3 \\
    S+L+L & 8.10  & 19.6  & 42.8  & 54.5  & 7.90  & 20.7  & 43.8  & 55.0 \\
    S+T+T & 15.90 & 13.2  & 31.8  & 42.8  & 15.25 & 13.9  & 32.9  & 43.6 \\
    G+L+L~\cite{chen2018} & 7.50  & 20.7  & 44.7  & 56.2  & \textbf{7.00} & 21.9  & \textbf{45.4} & \textbf{56.7} \\
    G+T+L & 7.60  & 20.7  & 44.3  & 55.9  & \textbf{7.00} & 21.8  & \textbf{45.4} & 56.6 \\
    G+T+T & 7.50  & \textbf{20.9} & 44.5  & 56.0  & \textbf{7.00} & \textbf{22.0} & 45.2  & 56.6 \\
    T+L+L & \textbf{7.30} & \textbf{20.9} & \textbf{44.8} & \textbf{56.3} & \textbf{7.00} & 21.9  & 45.3  & 56.6 \\
    T+L+T & \textbf{7.30} & 20.7  & 44.7  & 56.2  & \textbf{7.00} & 21.9  & 45.2  & 56.6 \\
    T+T+L & 7.50  & 20.8  & 44.3  & \textbf{56.3} & 7.10  & 21.7  & 45.0  & 56.5 \\
    T+T+T & \textbf{7.30} & \textbf{20.9} & 44.6  & 56.1  & \textbf{7.00} & 21.8  & 45.2  & 56.5 \\
    \bottomrule
     \end{tabular}%
  \label{tab:xretrieval_sup}%
 \end{table*}%
 
 \begin{figure*}[!ht]
\centering
\includegraphics[width=0.5\textwidth]{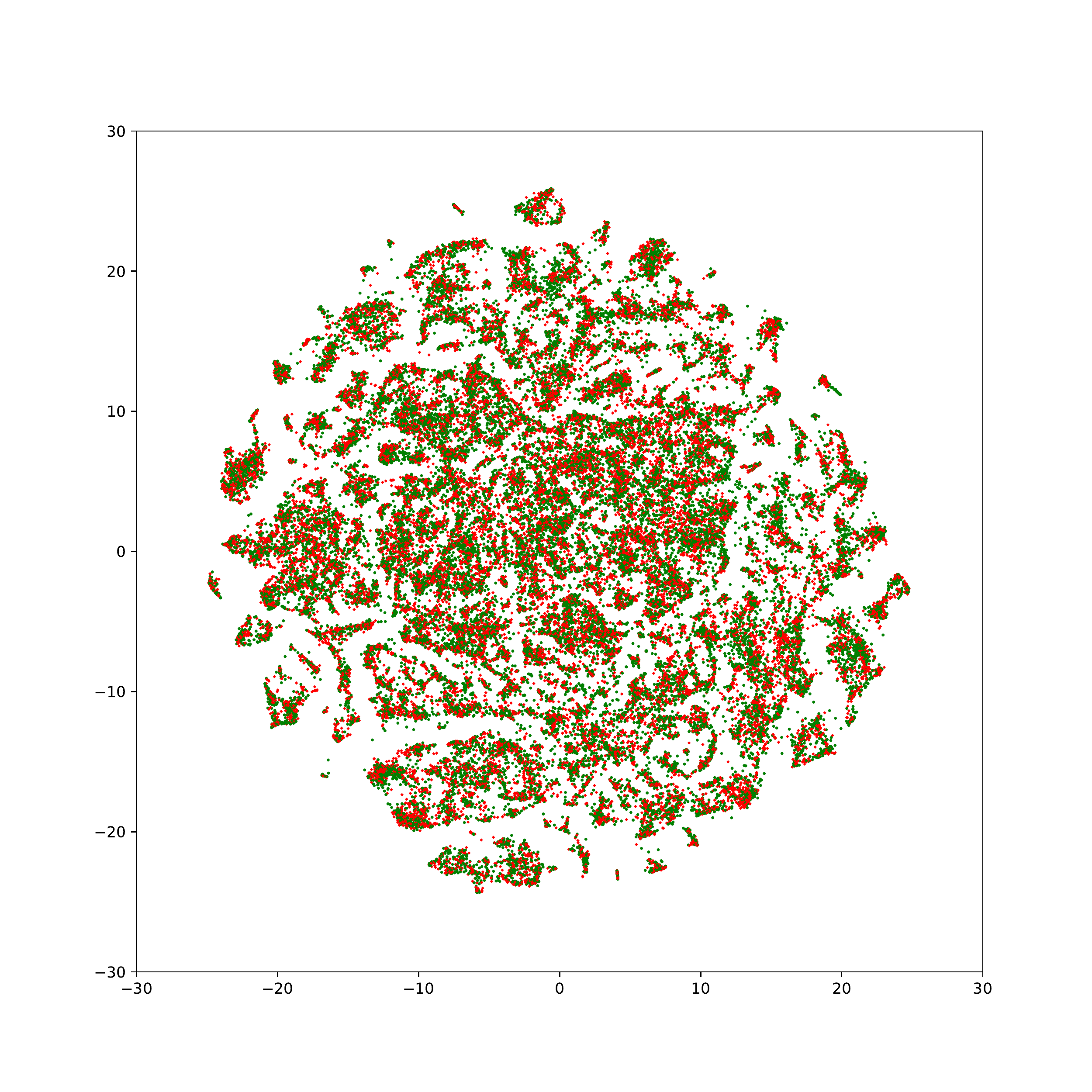}
\caption{t-SNE visualization of the visual (green) and textual (red) embeddings in the shared space. The two modalities' distributions are well aligned in the latent space.}
\label{fig:tsne}
\end{figure*}
 
 \begin{figure*}[!ht]
    \centering
    \includegraphics[width=\textwidth]{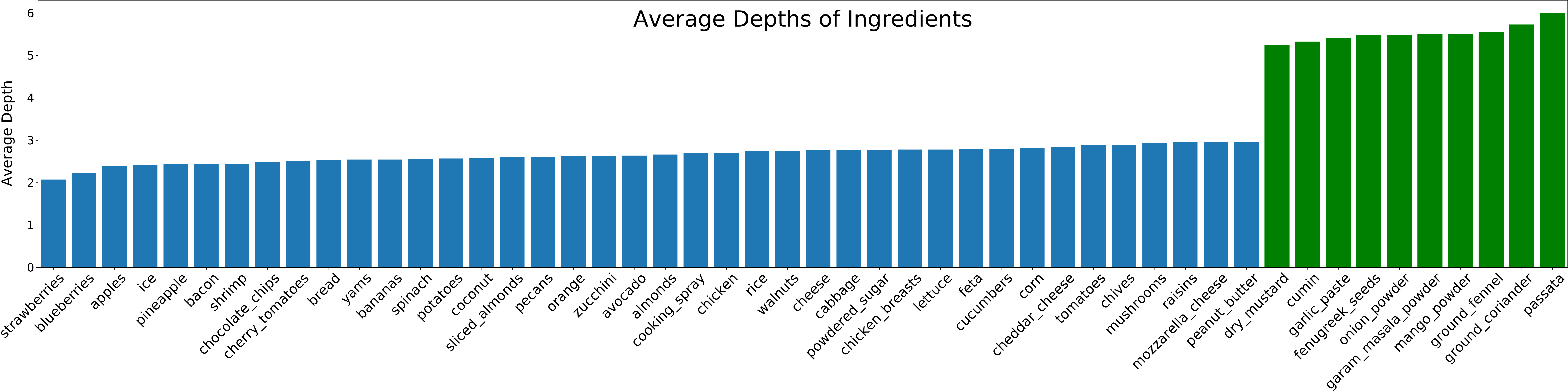}
    \caption{Average depths of the 40 shallowest (blue) and 10 deepest (green) ingredients in the test set.}
    \label{fig:ingr_depth_plot}
\end{figure*}

\begin{figure*}[!ht]
    \centering
    \subfloat[Broccoli rabe with white beans and parmesan]{
        \resizebox{0.7\textwidth}{!}{
            \Tree [.(-7-)
                [.(-6-)
                  \textcolor{red}{broccoli\_rabe}
                  [.(-3-) olive\_oil [.(-2-) dried\_chilies garlic ] ] ]
                [.(-5-)
                  [.(-4-) [.(-1-) salt black\_pepper ] white\_beans ]
                  parmigiano ] ]
        }
        \label{fig:ingr_tree_b}
    } \\
    \subfloat[Speedy $<$UNK$>$ quesadillas]{
        \resizebox{0.7\textwidth}{!}{
            \Tree [.(-5-)
                [.(-2-) chili\_sauce [.(-1-) miracle\_whip chili\_powder ] ]
                [.(-4-)
                  [.(-3-) flour\_tortillas \textcolor{red}{chicken\_breasts} ]
                  mexican\_cheese ] ]
        }
        \label{fig:ingr_tree_d}
    }
    \caption{Encoded ingredient hierarchical structures. Main ingredients are marked in red.}
    \label{fig:ingr_tree}
\end{figure*}

\begin{figure*}[!ht]
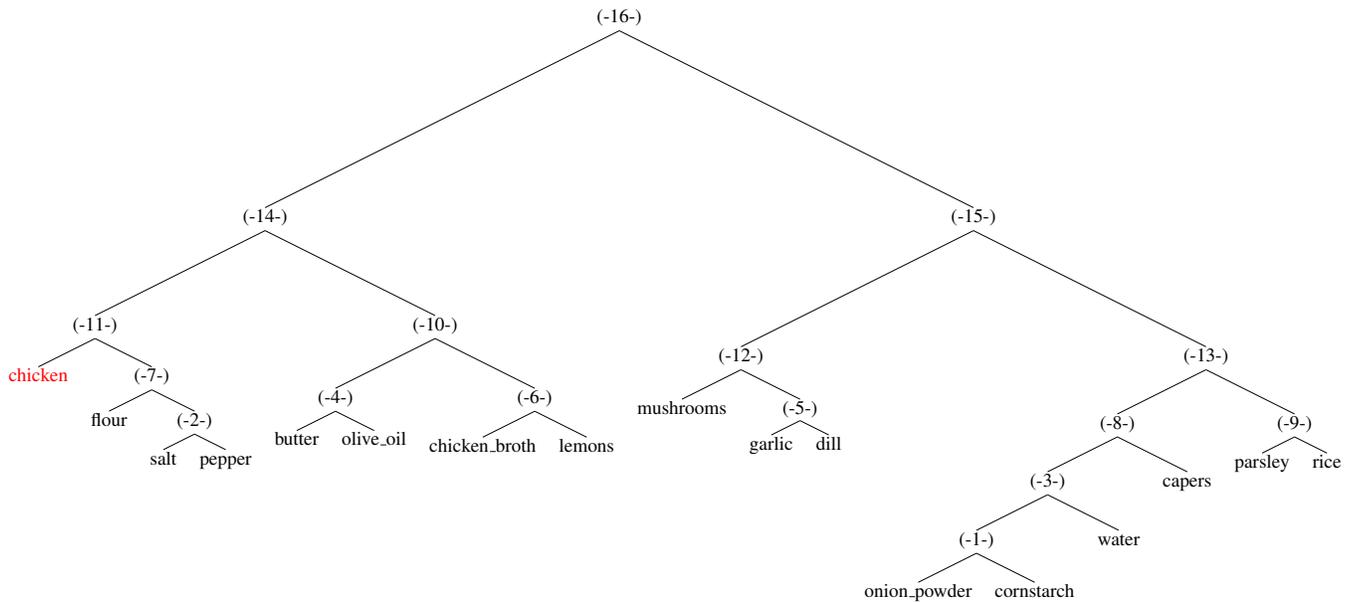
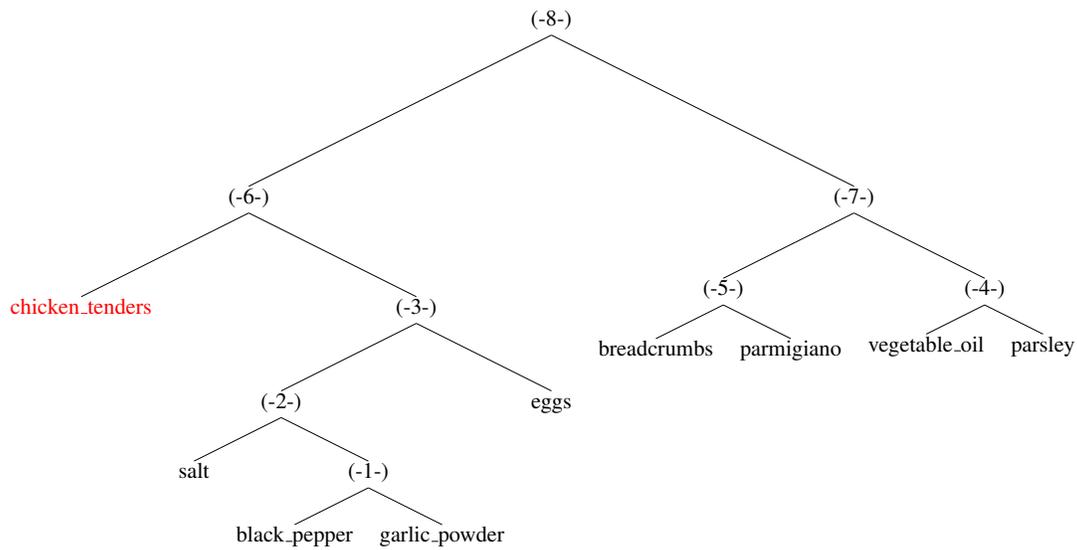

\vspace{-1em}
    \centering
    \subfloat[Crockpot chicken piccata]{
        \resizebox{\textwidth}{!}{
            \Tree [.(-16-)
                [.(-14-)
                  [.(-11-) \textcolor{red}{chicken} [.(-7-) flour [.(-2-) salt pepper ] ] ]
                  [.(-10-)
                    [.(-4-) butter olive\_oil ]
                    [.(-6-) chicken\_broth lemons ] ] ]
                [.(-15-)
                  [.(-12-) mushrooms [.(-5-) garlic dill ] ]
                  [.(-13-)
                    [.(-8-)
                      [.(-3-) [.(-1-) onion\_powder cornstarch ] water ]
                      capers ]
                    [.(-9-) parsley rice ] ] ] ]
        }
        \label{fig:ingr_tree_2_a}
    } \\
    \subfloat[Parmesan panko chicken tenders]{
        \resizebox{0.8\textwidth}{!}{
            \Tree [.(-8-)
                [.(-6-)
                  \textcolor{red}{chicken\_tenders}
                  [.(-3-)
                    [.(-2-) salt [.(-1-) black\_pepper garlic\_powder ] ]
                    eggs ] ]
                [.(-7-)
                  [.(-5-) breadcrumbs parmigiano ]
                  [.(-4-) vegetable\_oil parsley ] ] ]
        }
        \label{fig:ingr_tree_2_b}
    }
    \caption{More complex ingredient hierarchical structures.}
    \label{fig:ingr_tree_2}
\end{figure*}

Tab.~\ref{tab:xretrieval_sup} shows the complete cross-modal retrieval scores of all models evaluated in our paper. With the exception of \textbf{[S+T+T]}, all variants of our proposed framework outperform the state-of-the-art ACME~\cite{wang2019}. The performance of \textbf{[G+L+L]} (with attention) in text-to-image retrieval is the best, however, as we have shown in the ingredient detection and substitution experiments, it is unable to emphasize the main ingredients properly. Thus, the retrieval task may not fully  exploit the information richness of ingredient embeddings, which incite us to conduct more experiments beyond retrieval to measure the effectiveness of our proposed hierarchical recipe embeddings.

\subsubsection{The mysterious case of ACME}
In the main paper, we have explained that we retrained ACME on our modified data (i.e. creating canonical ingredients), and we did not get the same performance as reported in the original paper~\cite{wang2019}. We inspected the published code\footnote{https://github.com/hwang1996/ACME}, and discovered a coding bug, where the batch normalization layer after the last fully connected layer was not set in evaluation mode while testing, thus the test results are highly data-dependent. When we set up the experiments as exactly described by the authors, using original code, data and model provided by the authors, we can reproduce the results. However, after correcting the code, the evaluation performance decreases significantly. More specifically, we recorded the (medR; R@1, R@5 and R@10) scores on the 1k experiment as following: image-to-recipe: (1.55; 49.7, 79.1, 86.9), text-to-image: (1.20; 51.0, 79.6, 86.9). One possible explanation of why its performance on our modified dataset is not as good, is that we have removed the context from ingredient descriptions (e.g. ``chopped onion'' is now just ``onion'', thus ACME cannot exploit more information from ingredients as in the original work.

\subsubsection{Shared space visualization}
Fig.~\ref{fig:tsne} illustrates the t-SNE visualization of the two modalities' embeddings estimated by the \textbf{[T+T+L]} model. It can be observed that the two distributions are closely aligned in the shared latent space.

\subsection{Main Ingredient Detection}
\label{sec:main_ingr}

\begin{table*}[!ht]
  \centering
  \caption{Cross-modal retrieval performance after ingredient pruning. This table shows results of \textbf{[T+T+L]} after removing the last K ingredients, or only keeping the top K ingredients, or keeping ingredients up to depth of K.}
    \begin{tabular}{ccccccccc}
    \toprule
          & \multicolumn{4}{c}{\textbf{Image-to-Recipe}} & \multicolumn{4}{c}{\textbf{Recipe-to-Image}} \\
    \midrule
    \textbf{K} & \textbf{medR} & \textbf{R@1} & \textbf{R@5} & \textbf{R@10} & \textbf{medR} & \textbf{R@1} & \textbf{R@5} & \textbf{R@10} \\
    \midrule
    Original & 1.75  & 49.0  & 78.8  & 85.9  & 1.60  & 49.7  & 78.9  & 86.6 \\
    \midrule
    \multicolumn{9}{c}{Remove last K} \\
    \midrule
    1     & \textbf{1.10} & \textbf{51.1} & \textbf{79.9} & \textbf{87.0} & \textbf{1.00} & \textbf{51.9} & \textbf{79.9} & 86.8 \\
    2     & 1.50  & 0.5   & 79.4  & 86.7  & 1.20  & 50.8  & 79.3  & \textbf{87.1} \\
    3     & 1.85  & 48.7  & 78.2  & 85.9  & 1.80  & 49.4  & 77.8  & 86.0 \\
    4     & 1.90  & 46.3  & 76.6  & 84.7  & 1.95  & 47.8  & 76.6  & 84.9 \\
    \midrule
    \multicolumn{9}{c}{Keep first K} \\
    \midrule
    3     & 2.00  & 41.1  & 71.4  & 80.2  & 2.00  & 40.4  & 70.1  & 79.5 \\
    4     & 2.00  & 40.9  & 70.8  & 79.7  & 2.00  & 42.0  & 70.6  & 79.3 \\
    5     & 2.00  & 42.9  & 73.6  & 82.5  & 2.00  & 43.9  & 73.3  & 82.5 \\
    6     & 2.00  & \textbf{44.7} & \textbf{74.3} & \textbf{83.6} & 2.00  & \textbf{45.7} & \textbf{74.7} & \textbf{83.9} \\
    \midrule
    \multicolumn{9}{c}{Keep depth of K} \\
    \midrule
    3     & 2.00  & 44.8  & 75.1  & 83.3  & 2.00  & 45.1  & 74.4  & 83.1 \\
    4     & \textbf{1.70} & \textbf{49.7} & \textbf{78.5} & \textbf{86.2} & \textbf{1.50} & \textbf{49.9} & \textbf{78.7} & \textbf{85.9} \\
    \bottomrule
    \end{tabular}%
  \label{tab:ingr_prune_sup}%
\end{table*}%

Fig.~\ref{fig:ingr_depth_plot} illustrates the average depths of 50 ingredients, including 40 ingredients with lowest average depths, and 10 with highest average depths.
Fig.~\ref{fig:ingr_tree} and \ref{fig:ingr_tree_2} show more examples of ingredient tree structures generated by the ingredient encoder of \textbf{[T+T+L]}, the overall best performing model in our experiments. Three out of four recipes contain ``chicken'' as the main ingredient, with the exception of ``broccoli rabe'', in Fig.~\ref{fig:ingr_tree_b}. In these recipes, the ingredient encoder is able to impose the significance of main ingredients, but it fails to capture the main ingredient of ``Speedy $<$UNK$>$ quesadillas''\footnote{The title of this recipe contains a rarely used word that was left out of the vocabulary.}. 
This observation explains why the adaptation from this recipe, demonstrated in Fig.~\ref{fig:ingr_sub_fail}, does not successfully transform it into a new base. We will revisit the recipe in Fig.~\ref{fig:ingr_tree_d} in the next section.

\subsubsection{Ingredient Pruning}
\begin{figure*}[!ht]
    \centering
    \includegraphics[width=\textwidth]{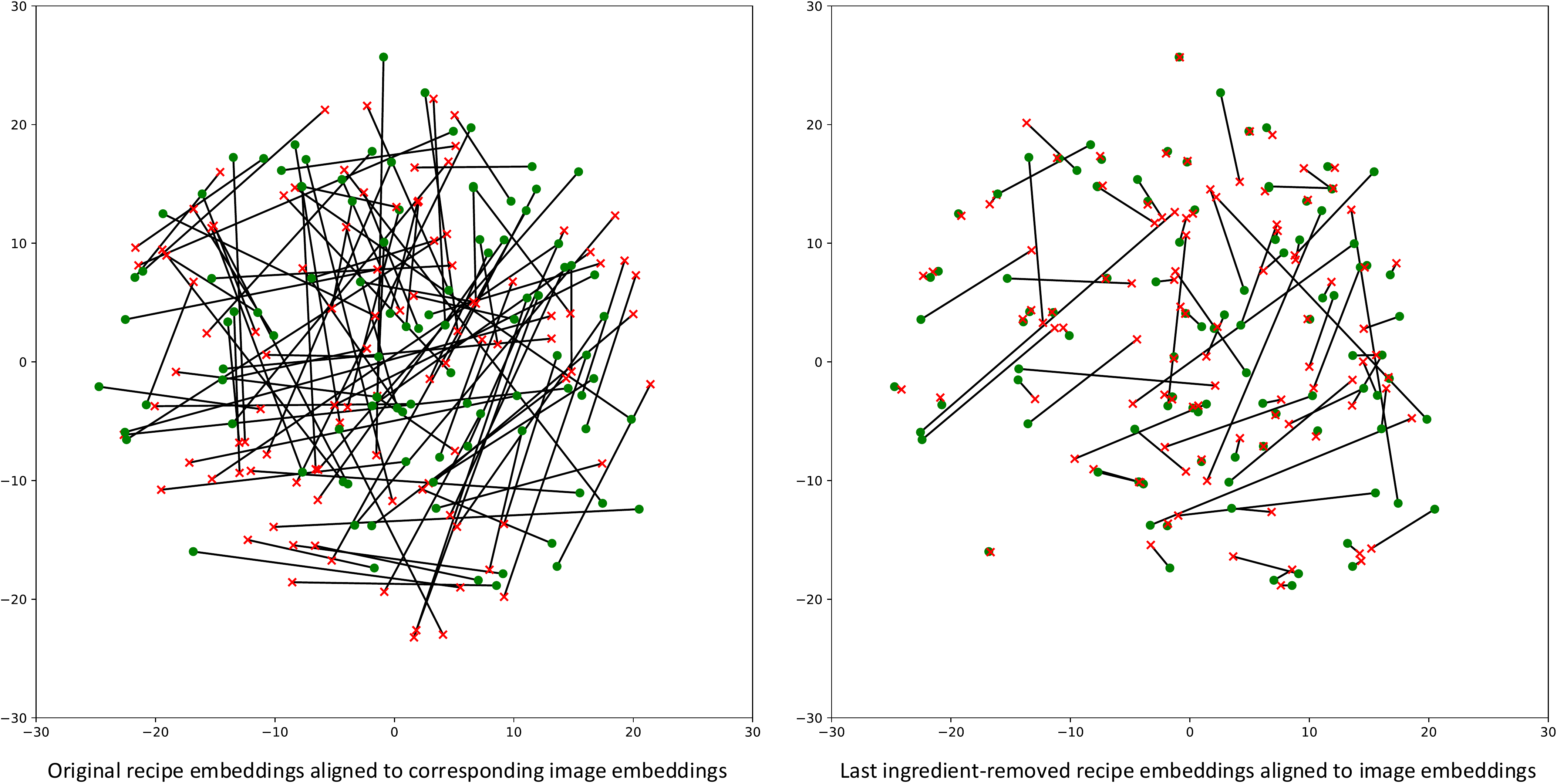}
    \caption{t-SNE visualization of the visual embeddings (green) and textual embeddings (red - left figure) of 100 samples, and the corresponding last-ingredient-removed recipe embeddings (red - right figure). Overall, removing the least important ingredient has effectively improve the cross-modal alignment, indicated by shorter distance between the image and recipe embeddings of the sample pair.}
    \label{fig:tsne_remove1}
\end{figure*}

Tab.~\ref{tab:ingr_prune_sup} shows the retrieval results of \textbf{[T+T+L]} after pruning ingredients. In addition to the removing K ingredients results in the main paper, we include results after removing all ingredients except the top K ingredients or ingredients up to the depth of K. These experiments show that the removal of many ingredients indeed leads to performance degradation, demonstrated in the ``Keep first K'' experiments. However, if a reasonable number of ingredients is retained, as shown in ``Remove last K'' and ``Keep depth of K'' experiments, the retrieval performance actually improves. Fig.~\ref{fig:tsne_remove1} shows examples where removing the least important ingredients has improved the cross-modal alignment between image and text embeddings in the latent space, without learning a new model.

\subsection{Ingredient Substitution}
\label{sec:ingr_sub_sup}

Tab.~\ref{tab:subs_from_beef} show the Success Rates of replacing ``beef'' with other ingredients, carried out in the same way as the substitution experiments in the main text. There are mixed results in these experiments. pic2rec performs better in the substitution from ``beef'' to ``chicken''. These results are different from the observations of replacing ``chicken'' and ``pork'' in the main text. We hypothesize that the original ``beef'' recipes are influenced more by other ingredients, thus the simple ingredient encoders in pic2rec, ACME, as well as [\textbf{S+L+L}], where all ingredients are considered equally important, actually perform better in this task.

\begin{table*}[!ht]
  \centering
  \small
  \caption{Success rates of substitution from ``beef'' to other ingredients. Best results are marked in bold.}
    \begin{tabular}{lccccccccc}
    \toprule
           & \multicolumn{2}{c}{\textbf{To Chicken}} & \multicolumn{2}{c}{\textbf{To Apple}} & \multicolumn{2}{c}{\textbf{To Pork}} & \multicolumn{2}{c}{\textbf{To Fish}} &  \\
    \cmidrule(lr){1-1} \cmidrule(lr){2-3} \cmidrule(lr){4-5} \cmidrule(lr){6-7} \cmidrule(lr){8-9} \cmidrule(lr){10-10}
    \textbf{Methods} & \textbf{R-2-I} & \textbf{R-2-R} & \textbf{R-2-I} & \textbf{R-2-R} & \textbf{R-2-I} & \textbf{R-2-R} & \textbf{R-2-I} & \textbf{R-2-R} & \textbf{medR} \\
    \cmidrule(lr){1-1} \cmidrule(lr){2-3} \cmidrule(lr){4-5} \cmidrule(lr){6-7} \cmidrule(lr){8-9} \cmidrule(lr){10-10}
    pic2rec~\cite{salvador2017}    & \textbf{28.3} & \textbf{32.0} & 4.0    & \textbf{6.3}    & 10.7   & \textbf{14.2} & 1.8    & 1.2    & 5.5 \\
    ACME~\cite{wang2019}   & 18.8   & 15.7   & 3.8 & 3.5 & 8.6   & 10.5   & 2.1    & 1.7    & 10.5 \\
    S+L+L & 24.8   & 12.5   & \textbf{7.1}    & 4.8    & 11.9   & 5.9    & \textbf{3.0} & \textbf{3.0} & 4.5 \\
    S+T+T & 26.7   & 8.6    & 6.2    & 4.1    & 11.9   & 4.7    & 2.4    & 2.1    & 7.0 \\
    G+L+L~\cite{chen2018} & 25.7   & 12.8   & 6.0    & 4.0    & 11.6   & 5.4    & 2.4    & 2.7    & 7.0 \\
    G+T+L & 25.7   & 13.0   & 5.8    & 5.3    & \textbf{13.4} & 6.0    & 2.8    & 2.7    & 5.0 \\
    G+T+T & 26.1   & 12.0   & 6.1    & 5.1    & 11.5   & 5.8    & 2.7    & 2.6    & 5.5 \\
    T+L+L & 26.0   & 14.3   & 5.9    & 4.7    & 12.0   & 6.1    & 1.9    & 2.4    & 7.0 \\
    T+L+T & 28.0   & 14.2   & 6.6    & 5.5    & 11.4   & 6.0    & 2.6    & 2.5    & 4.5 \\
    T+T+L & 26.1   & 13.7   & 5.9    & 4.8    & 11.7   & 6.1    & 2.4    & 2.9    & 5.0 \\
    T+T+T & 26.4   & 14.7   & 6.2    & 5.4    & 11.5   & 6.1    & 2.2    & 2.7    & \textbf{3.0} \\
    \bottomrule
    \end{tabular}%
  \label{tab:subs_from_beef}%
\end{table*}%

\begin{figure*}[!ht]
    \centering
    \subfloat[]{
        \resizebox{\textwidth}{!}{
            \includegraphics[trim=1 1 1 1,clip]{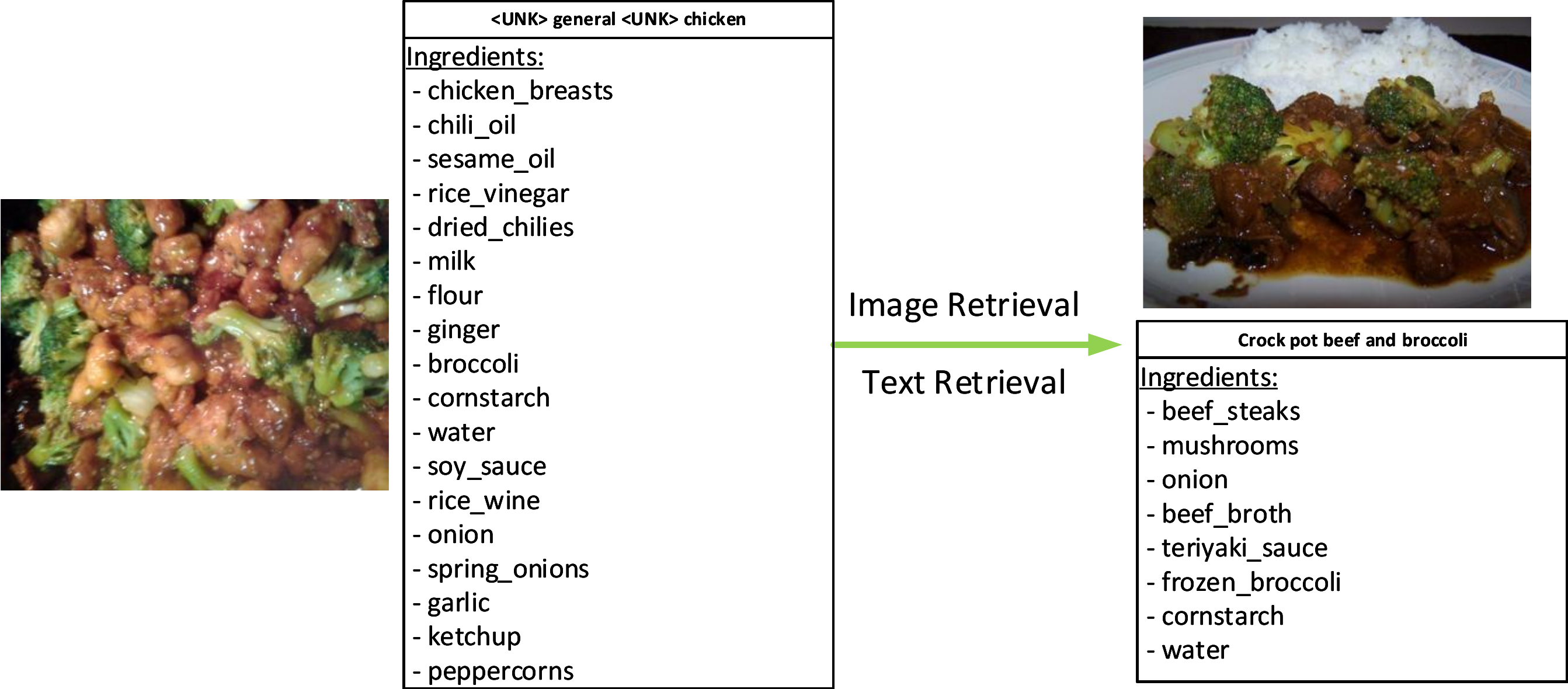}
        }
        \label{fig:c2b_success_a}
    } \\
    \subfloat[]{
        \resizebox{\textwidth}{!}{
            \includegraphics[trim=1 1 1 1,clip]{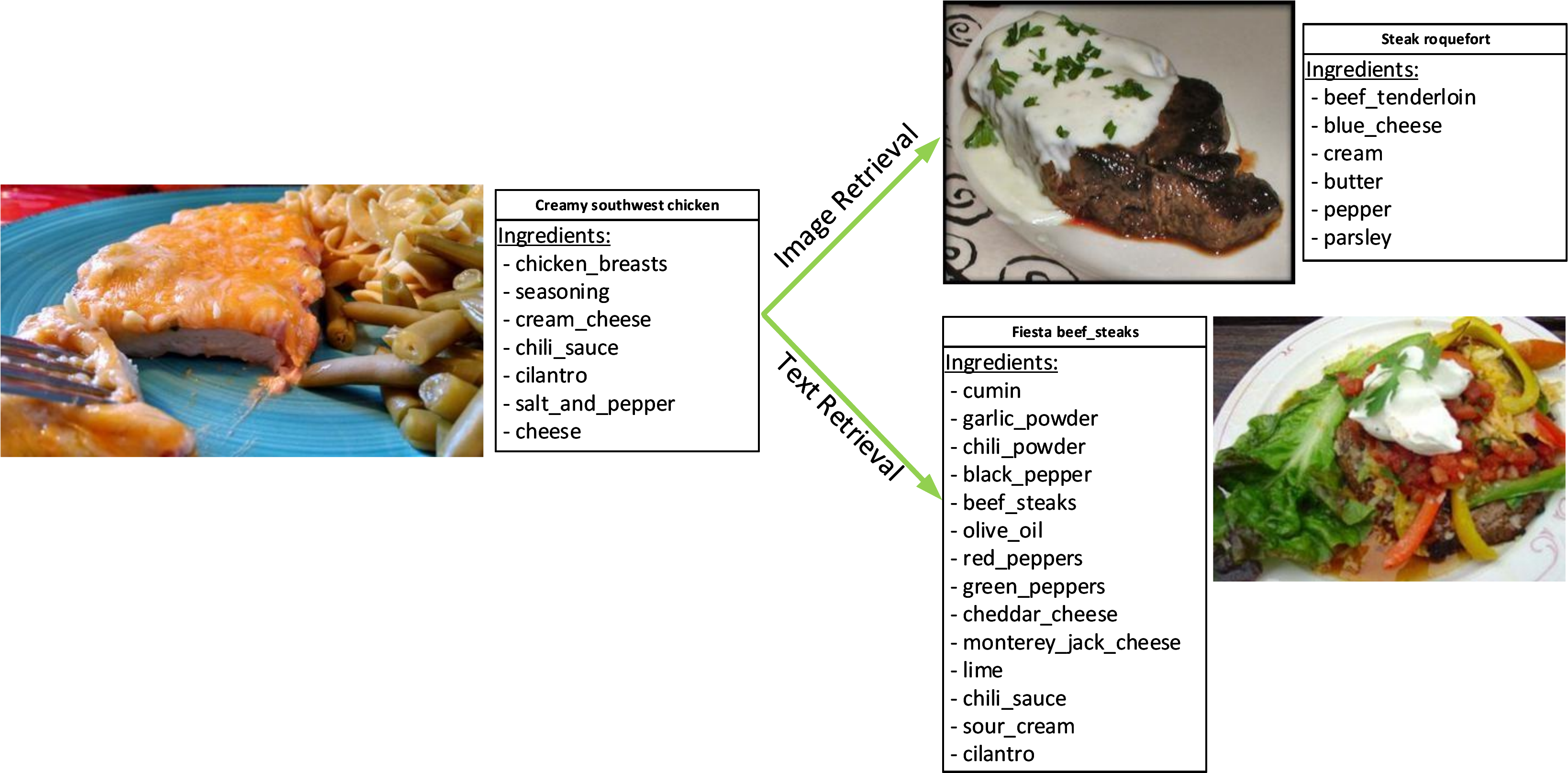}
        }
        \label{fig:c2b_success_b}
    }
    \caption{Ingredient substitutions from {\em chicken-to-beef}.}
    \label{fig:c2b_success}
\end{figure*}

\begin{figure*}[!ht]
    \centering
    \subfloat[]{
        \resizebox{\textwidth}{!}{
            \includegraphics[trim=1 1 1 1,clip]{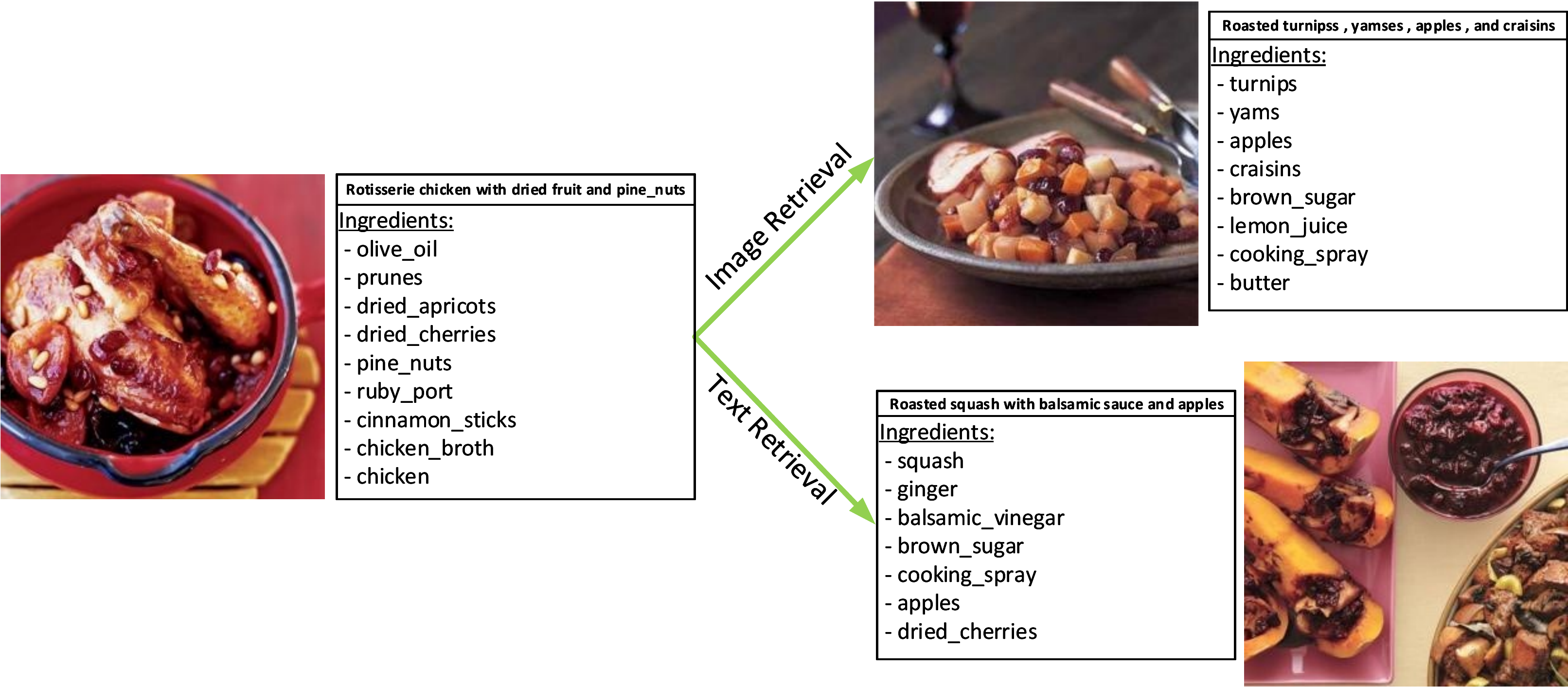}
        }
        \label{fig:c2a_success_a}
    } \\
    \subfloat[]{
        \resizebox{\textwidth}{!}{
            \includegraphics[trim=1 1 1 1,clip]{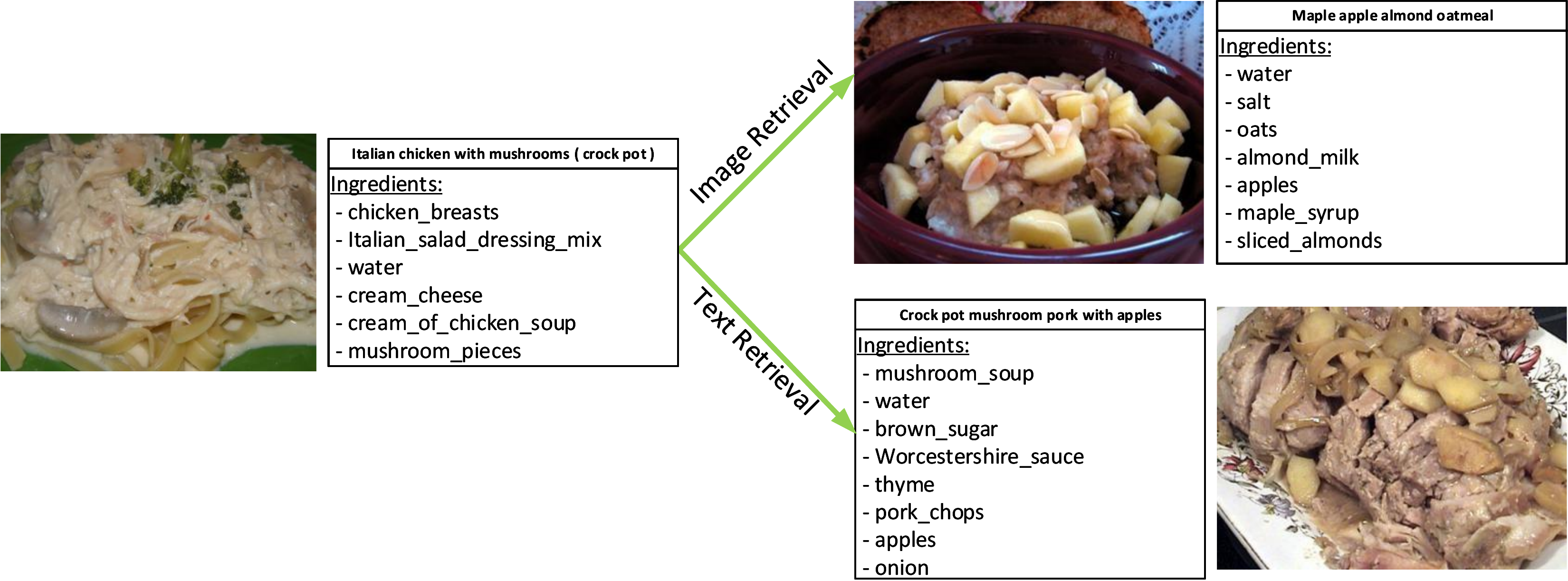}
        }
        \label{fig:c2a_success_b}
    }
    \caption{Ingredient substitutions from {\em chicken-to-apple}.}
    \label{fig:c2a_success}
\end{figure*}

\begin{figure*}[!ht]
    \centering
    \subfloat[Chicken-to-beef]{
        \resizebox{\textwidth}{!}{
            \includegraphics[trim=1 1 1 1,clip]{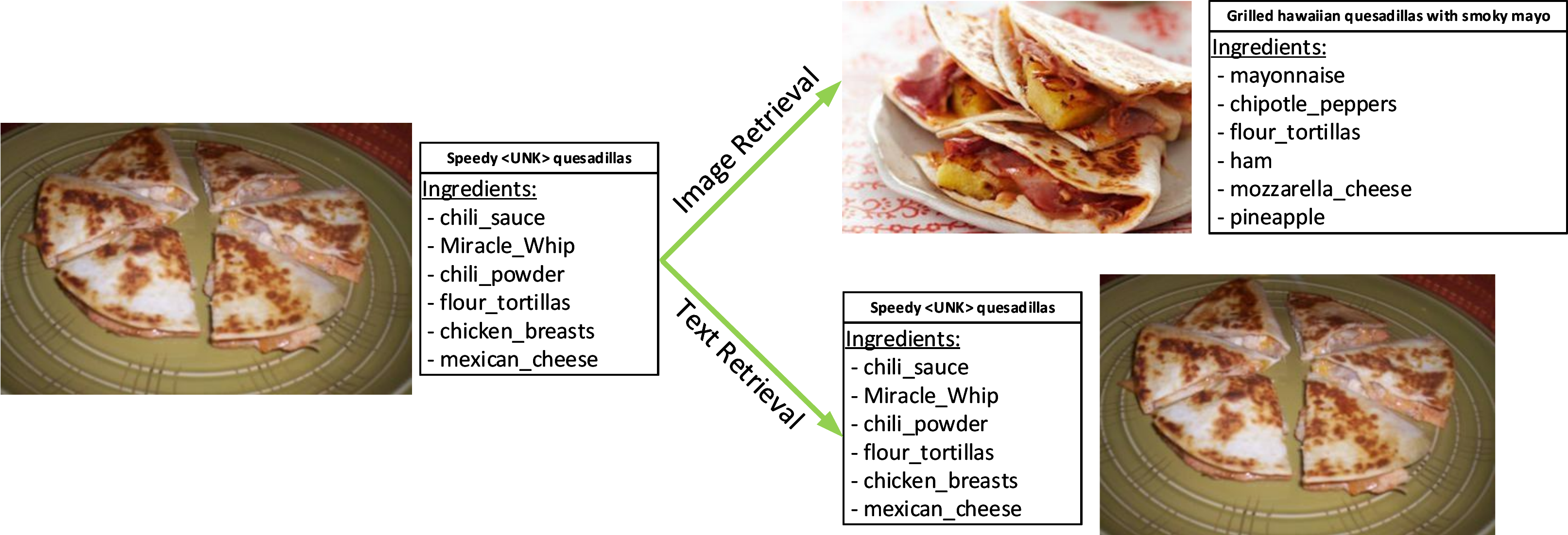}
        }
        \label{fig:c2b_fail_a}
    } \\
    \subfloat[Chicken-to-apple]{
        \resizebox{\textwidth}{!}{
            \includegraphics[trim=1 1 1 1,clip]{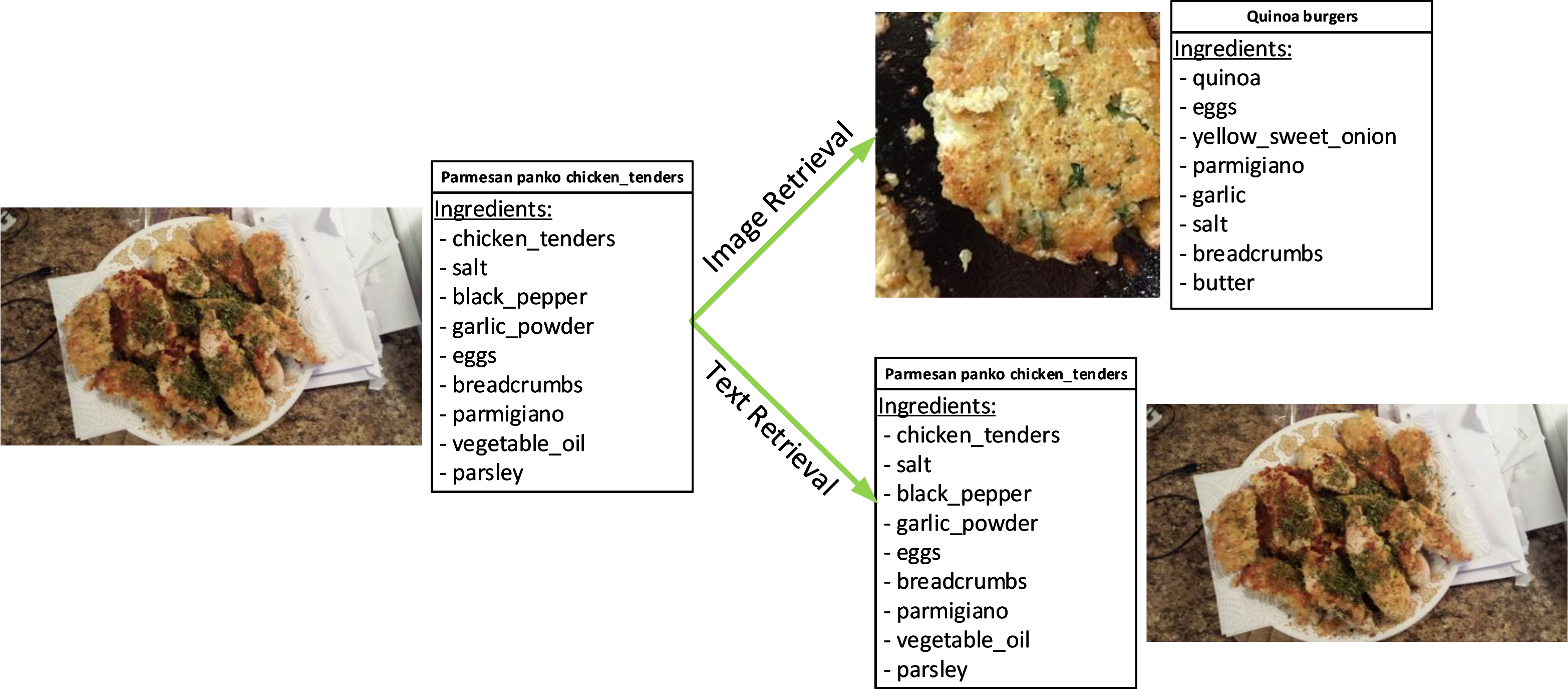}
        }
        \label{fig:c2a_fail_b}
    }
    \caption{Ingredient substitutions from {\em chicken} that fail to transform the recipes to other bases.}
    \label{fig:ingr_sub_fail}
\end{figure*}

Additional visual results of the ingredient substitution experiments described in the main paper, where we replace ``chicken'' in the test recipes with ``beef'' or ``apple'', are presented in Fig.~\ref{fig:c2b_success}-\ref{fig:ingr_sub_fail}.

\begin{figure*}[!ht]
    \centering
    \resizebox{\textwidth}{!}{
        \includegraphics[trim=1 1 1 1,clip]{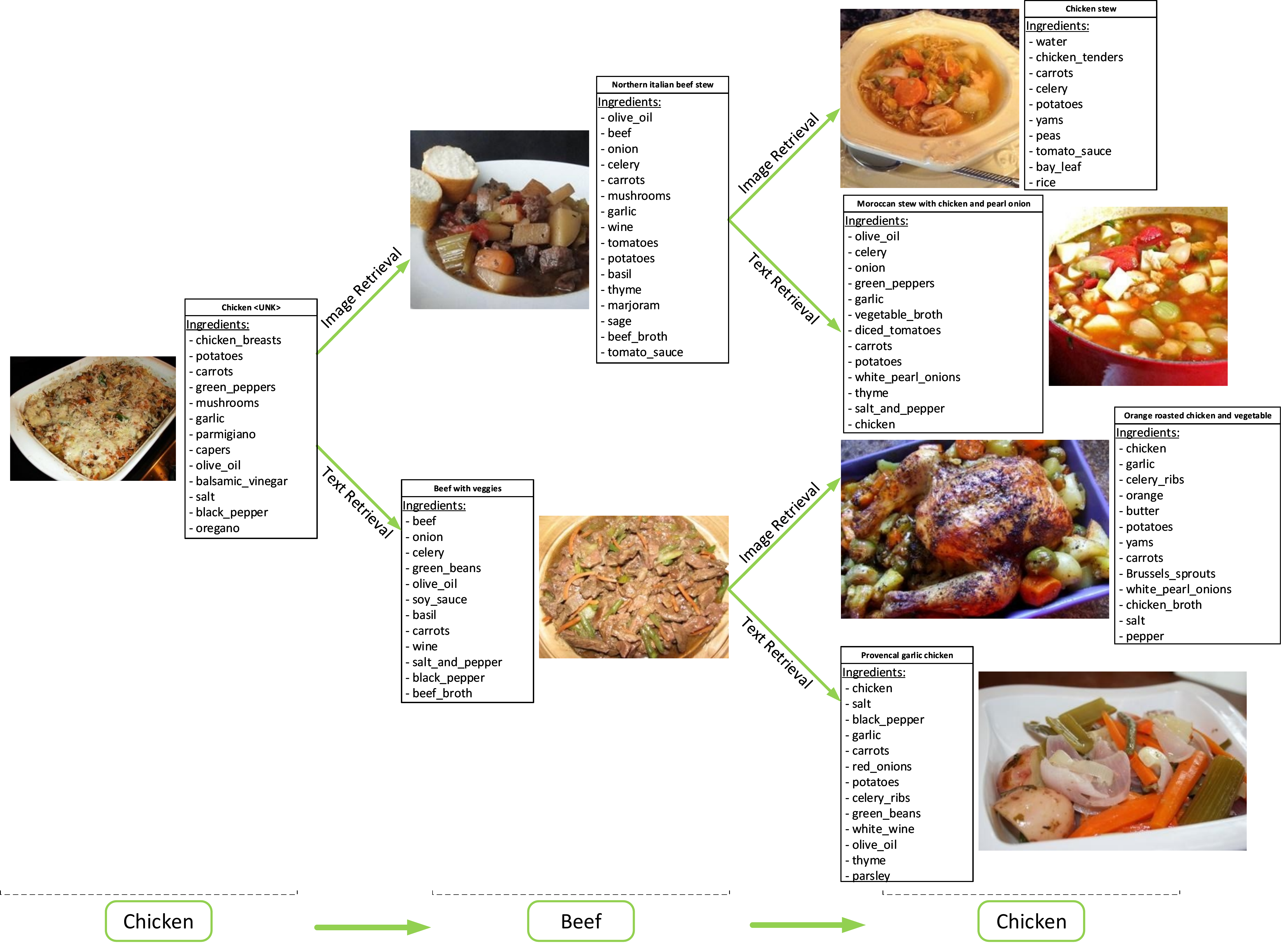}
    }
    \caption{Ingredient substitutions from {\em chicken-to-beef-to-chicken}. The ``chicken'' recipe on the left hand side is first adapted to ``beef'' to retrieve two top-1 recipes based on image and text embeddings, respectively, in the middle. These two recipes are further adapted to replace ``beef'' with ``chicken'' to retrieve four recipes displayed on the right hand side.}
    \label{fig:c2b2c_success}
\end{figure*}

\begin{figure*}[!ht]
    \centering
    \resizebox{0.95\textwidth}{!}{
        \includegraphics[trim=1 1 1 1,clip]{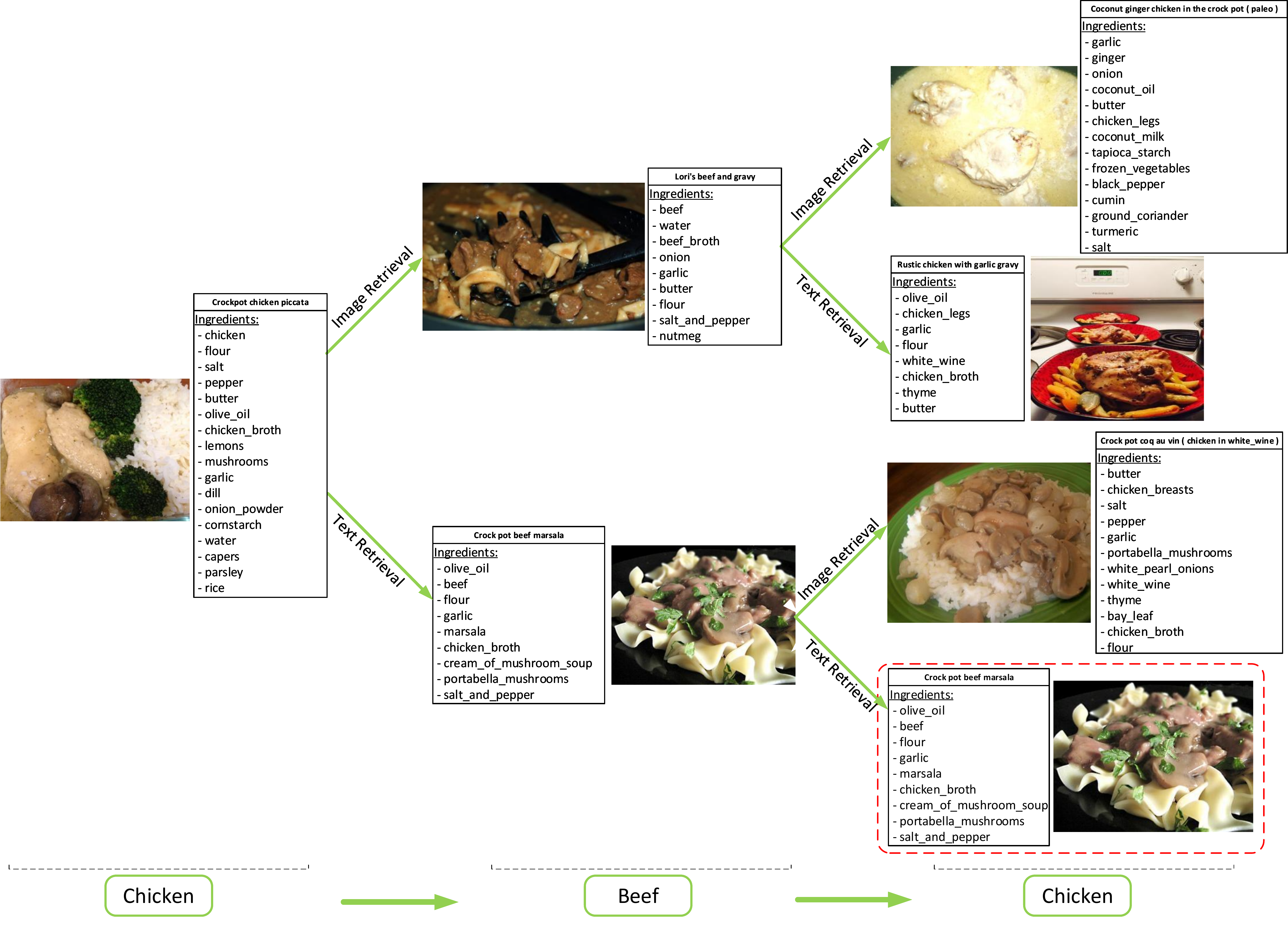}
    }
    \caption{Ingredient substitutions from {\em chicken-to-beef-to-chicken}. The last adaptation, highlighted in red box, is unsuccessful. The retrieved text embedding belongs to the same recipe before substitution.}
    \label{fig:c2b2c_one_fail}
\end{figure*}

Fig.~\ref{fig:c2b_success_a} illustrates that 
after replacing the main ingredient and performing retrieval of either image or text embeddings, we receive the same recipe. In this particular case the image and recipe encoders of the \textbf{[T+T+L]} model have generated a very well matched pair of embeddings of ``Crock pot beef and broccoli'', and the embedding of the modified recipe lies close to them in the shared latent space. 
Fig.~\ref{fig:c2b_success_b} shows another interesting example. The original recipe contains a lot of cream and cheese, and the retrieved recipes not only have ``beef'' but also use cream and cheese heavily: either shown visually in the retrieved image, or in the text description.

Fig.~\ref{fig:c2a_success} demonstrates two successful adaptations from ``chicken'' to ``apple''. What is interesting in these examples is the similarity of the preparation processes in the original and the resulting recipes, while they have vastly different ingredient sets. In Fig.~\ref{fig:c2a_success_a}, the original dish has {\em roasted chicken}, and the retrieved dishes have {\em roasted apple}, and all images show a reddish color tone. It means that the encoders have learned the linking between the image features dominated by red pixel values, and ``roasting'' preparation technique, thus it is able to adapt how ``chicken'' is roasted, to cook ``apple'' in a similar way. The same phenomenon is observed in Fig.~\ref{fig:c2a_success_b}, in which the preparation method is {\em crock pot cooking}.

However, our recipe encoder does not succeed in all substitutions. Fig.~\ref{fig:ingr_sub_fail} shows two examples in which the recipe encoder fails to transform chicken-based recipes to other bases. Particularly, the results of text retrieval are the same as the originals. These observations can be explained by how ingredients are embedded for two recipes, as shown in Fig.~\ref{fig:ingr_tree_d} and Fig.~\ref{fig:ingr_tree_2_b} 
As the ingredient hierarchies of these two recipes show, ingredients contribute similarly to the final embedding based on their depths, hence the importance of ``chicken'' in these recipes is diminished. Consequently, replacing ``chicken'' with another ingredient was not enough to move the embedding for a substantial distance from its original coordinates in the latent space.

We conduct another set of experiments, in which the original recipes are first adapted from ``chicken'' to ``beef'', and the retrieved recipes are further transformed from ``beef'' back to ``chicken''. Some examples are illustrated in Fig.~\ref{fig:c2b2c_success}. In this figure, the first substitution results in vastly different dishes, but they share similarity in preparation that mostly includes {\em combine} and {\em stir}. In the second substitutions, {\em beef stew} is transformed into {\em chicken stew} (top-right), and {\em beef with veggies} is adapted to {\em chicken with veggies} (bottom-right). One failed substitution among such experiments is shown in Fig.~\ref{fig:c2b2c_one_fail}.

\subsection{Action Word Extraction}
\label{sec:action_word}

\begin{figure*}[!hb]
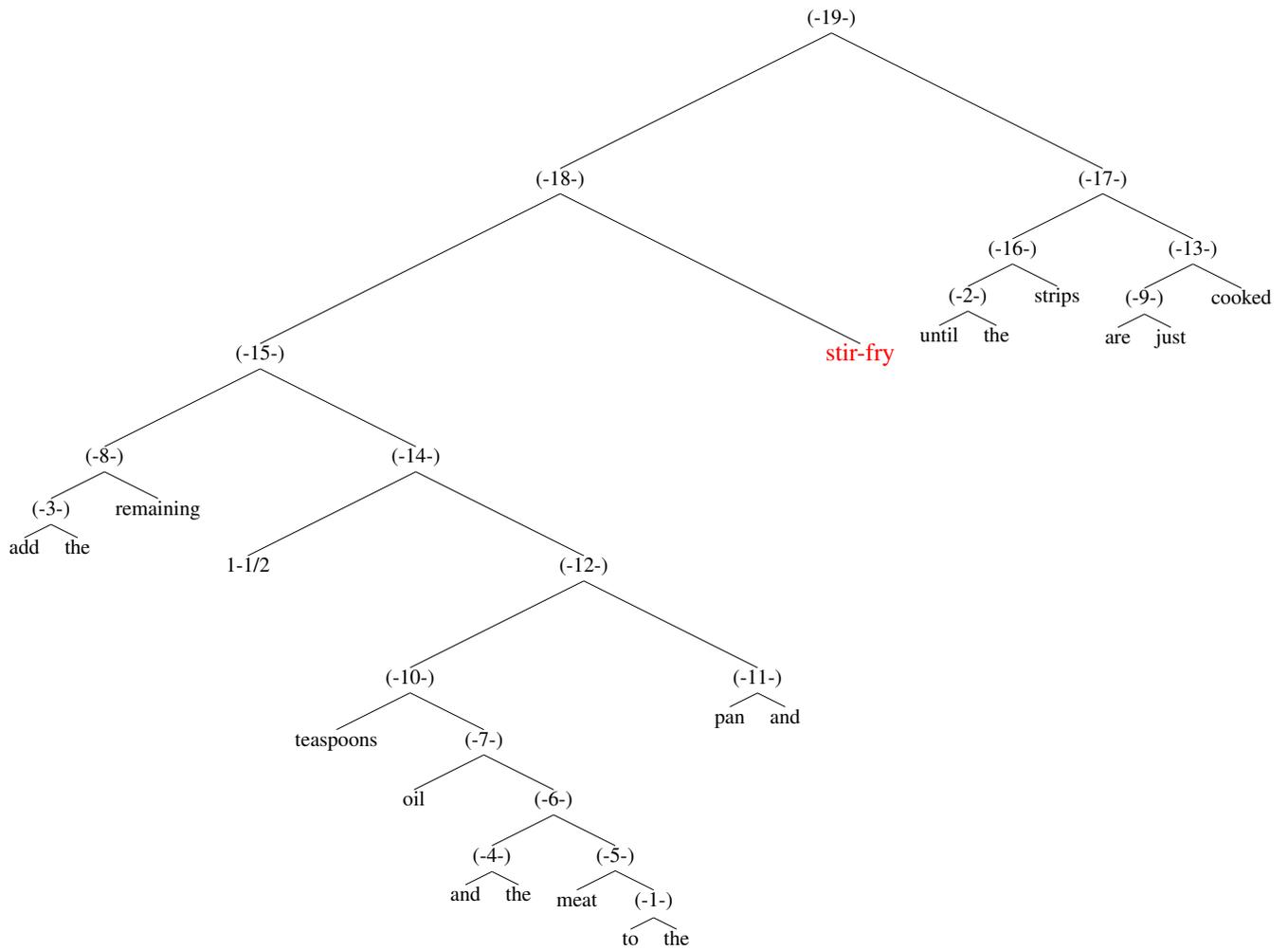

\vspace{-1em}
    \centering
    \resizebox{\textwidth}{!}{
        \Tree [.(-19-)
            [.(-18-)
              [.(-15-)
                [.(-8-) [.(-3-) add the ] remaining ]
                [.(-14-)
                  1-1/2
                  [.(-12-)
                    [.(-10-)
                      teaspoons
                      [.(-7-)
                        oil
                        [.(-6-)
                          [.(-4-) and the ]
                          [.(-5-) meat [.(-1-) to the ] ] ] ] ]
                    [.(-11-) pan and ] ] ] ]
              \textcolor{red}{\Large stir-fry} ]
            [.(-17-)
              [.(-16-) [.(-2-) until the ] strips ]
              [.(-13-) [.(-9-) are just ] cooked ] ] ]
        }
    \caption{Tree structure of a long sentence. It is the enlarged rendering of Fig. 6a in the main text.}
    \label{fig:sentence_tree_long}
\end{figure*}

\begin{figure*}[!ht]
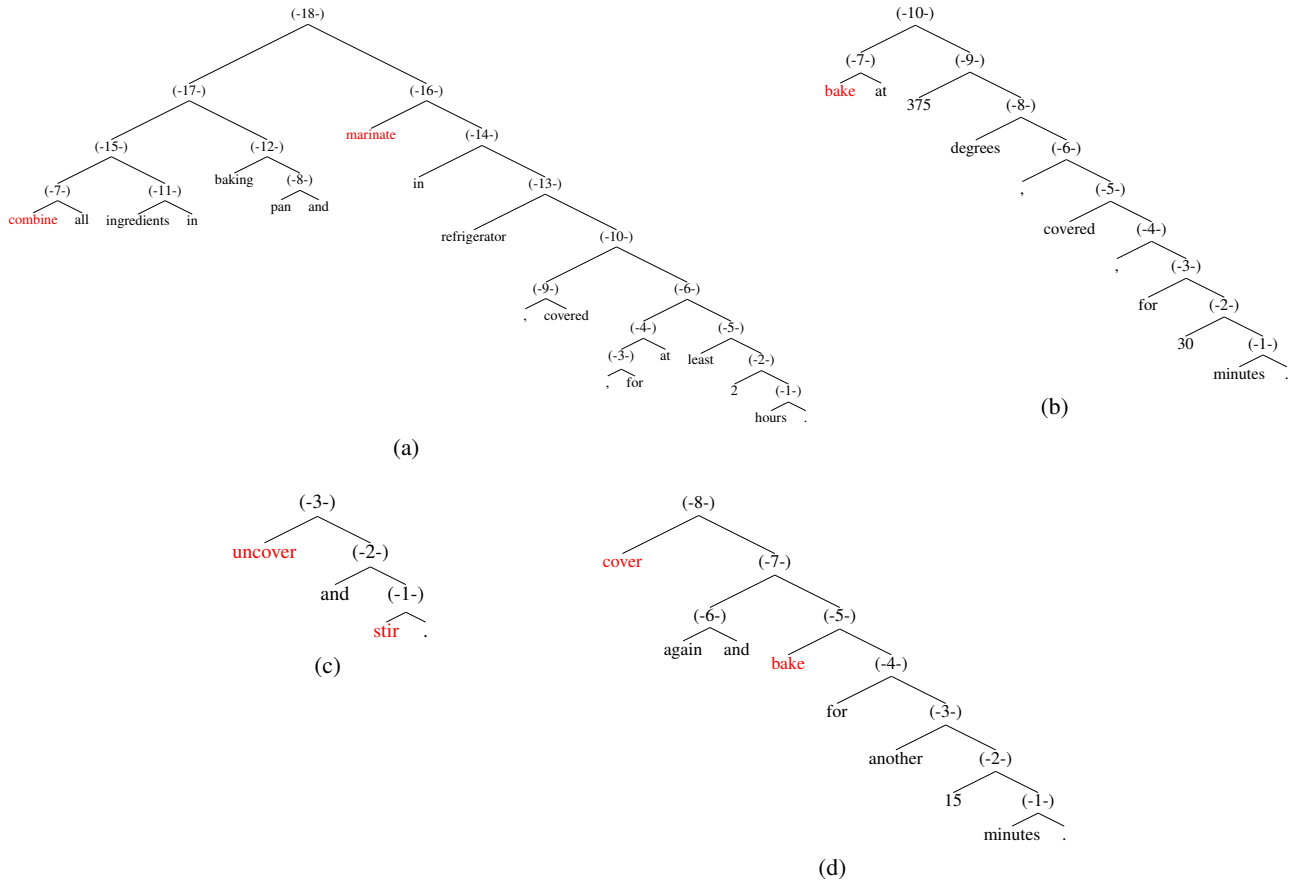

    \centering
    \subfloat[]{
        \resizebox{0.6\textwidth}{!}{
            \Tree [.(-18-)
            [.(-17-)
              [.(-15-) [.(-7-) \textcolor{red}{combine} all ] [.(-11-) ingredients in ] ]
              [.(-12-) baking [.(-8-) pan and ] ] ]
            [.(-16-)
              \textcolor{red}{marinate}
              [.(-14-)
                in
                [.(-13-)
                  refrigerator
                  [.(-10-)
                    [.(-9-) , covered ]
                    [.(-6-)
                      [.(-4-) [.(-3-) , for ] at ]
                      [.(-5-) least [.(-2-) 2 [.(-1-) hours . ] ] ] ] ] ] ] ] ]
        }
        \label{fig:sentence_tree_1_a}
    } 
    \subfloat[]{
        \resizebox{0.35\textwidth}{!}{
            \Tree [.(-10-)
            [.(-7-) \textcolor{red}{bake} at ]
            [.(-9-)
              375
              [.(-8-)
                degrees
                [.(-6-)
                  ,
                  [.(-5-)
                    covered
                    [.(-4-)
                      ,
                      [.(-3-) for [.(-2-) 30 [.(-1-) minutes . ] ] ] ] ] ] ] ] ]
        }
        \label{fig:sentence_tree_1_b}
    }
    \\
    \subfloat[]{
        \resizebox{0.15\textwidth}{!}{
            \Tree [.(-3-) \textcolor{red}{uncover} [.(-2-) and [.(-1-) \textcolor{red}{stir} . ] ] ]
        }
        \label{fig:sentence_tree_1_c}
    }
    \hspace{2cm}
    \subfloat[]{
        \resizebox{0.35\textwidth}{!}{
            \Tree [.(-8-)
        \textcolor{red}{cover}
        [.(-7-)
          [.(-6-) again and ]
          [.(-5-)
            \textcolor{red}{bake}
            [.(-4-)
              for
              [.(-3-) another [.(-2-) 15 [.(-1-) minutes . ] ] ] ] ] ] ]
        }
        \label{fig:sentence_tree_1_d}
    }
    \caption{Sentence trees of a recipe generated by \textbf{[T+T+L]}. Action words are manually highlighted in red. Some sentences contain more than one action word, however, as stated in the main paper, we only inspect the leaf closest to tree root.}
    \label{fig:sentence_tree_1}
\end{figure*}

\begin{figure*}[!ht]
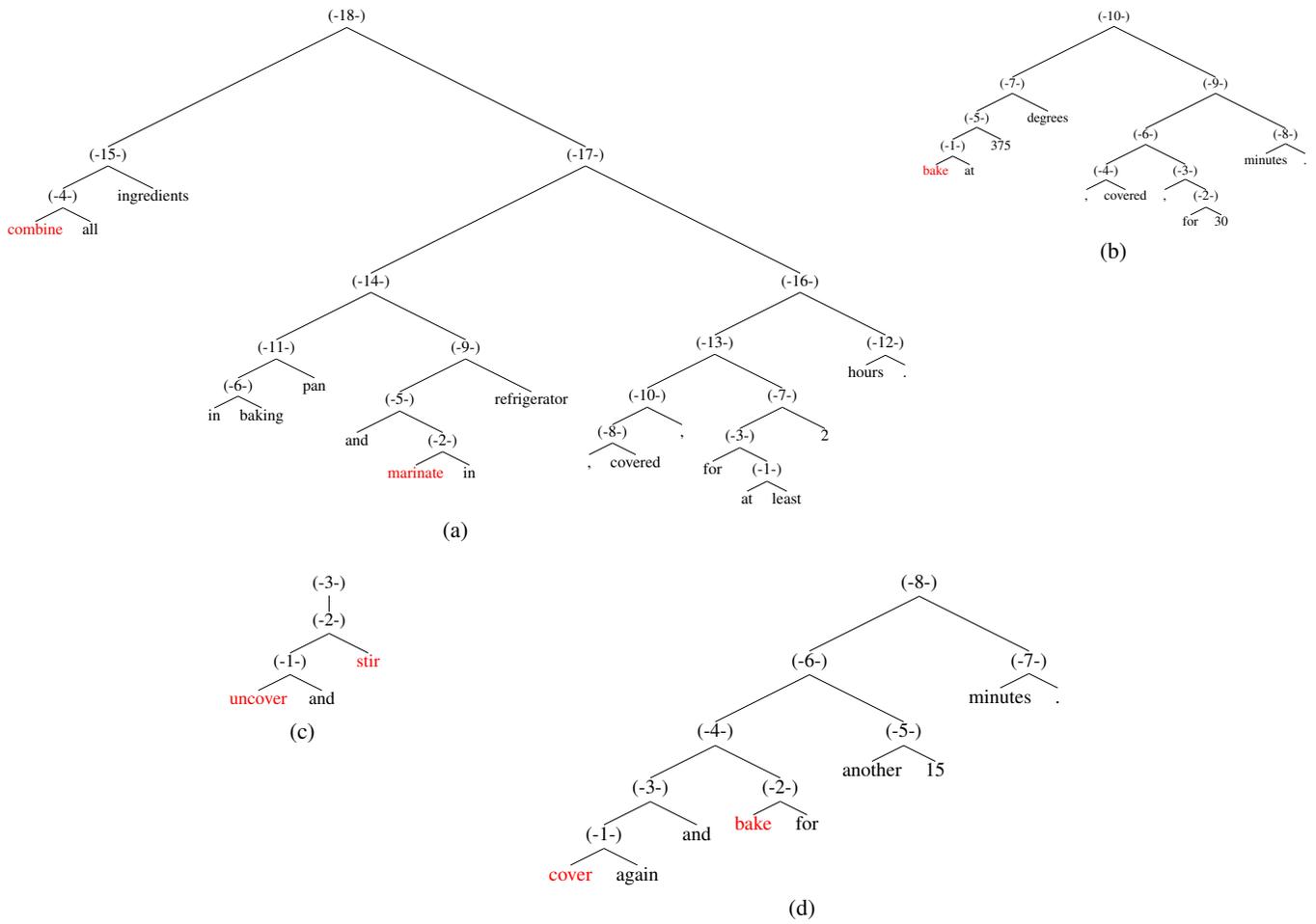

\vspace{-1em}
    \centering
    \subfloat[]{
        \resizebox{0.7\textwidth}{!}{
            \Tree [.(-18-)
            [.(-15-) [.(-4-) \textcolor{red}{combine} all ] ingredients ]
            [.(-17-)
              [.(-14-)
                [.(-11-) [.(-6-) in baking ] pan ]
                [.(-9-)
                  [.(-5-) and [.(-2-) \textcolor{red}{marinate} in ] ]
                  refrigerator ] ]
              [.(-16-)
                [.(-13-)
                  [.(-10-) [.(-8-) , covered ] , ]
                  [.(-7-) [.(-3-) for [.(-1-) at least ] ] 2 ] ]
                [.(-12-) hours . ] ] ] ]
        }
        \label{fig:sentence_tree_2_a}
    } 
    \subfloat[]{
        \resizebox{0.3\textwidth}{!}{
            \Tree [.(-10-)
            [.(-7-) [.(-5-) [.(-1-) \textcolor{red}{bake} at ] 375 ] degrees ]
            [.(-9-)
              [.(-6-) [.(-4-) , covered ] [.(-3-) , [.(-2-) for 30 ] ] ]
              [.(-8-) minutes . ] ] ]
        }
        \label{fig:sentence_tree_2_b}
    }
    \\
    \subfloat[]{
        \resizebox{0.12\textwidth}{!}{
            \Tree [.(-3-) [.(-2-) [.(-1-) \textcolor{red}{uncover} and ] \textcolor{red}{stir} ] . ]
        }
        \label{fig:sentence_tree_2_c}
    }
    \hspace{2cm}
    \subfloat[]{
        \resizebox{0.4\textwidth}{!}{
            \Tree [.(-8-)
            [.(-6-)
              [.(-4-)
                [.(-3-) [.(-1-) \textcolor{red}{cover} again ] and ]
                [.(-2-) \textcolor{red}{bake} for ] ]
              [.(-5-) another 15 ] ]
            [.(-7-) minutes . ] ]
        }
        \label{fig:sentence_tree_2_d}
    }
    \caption{Sentence trees of the same recipe in Fig.~\ref{fig:sentence_tree_1}, generated by \textbf{[T+T+T]}.}
    \label{fig:sentence_tree_2}
\end{figure*}

Fig.~\ref{fig:sentence_tree_long} illustrates the inferred structure of the long sentence.
Furthermore, we visualize the encoded sentence hierarchies of the same recipe by two models, \textbf{[T+T+L]} and \textbf{[T+T+T]}, in Fig.~\ref{fig:sentence_tree_1} and \ref{fig:sentence_tree_2}, respectively. These examples demonstrate that the sentence encoder of \textbf{[T+T+L]} model has learned to emphasize the action words consistently across different sentences, as evidenced in Fig.~\ref{fig:sentence_tree_1}. 
Whereas, the sentence encoder of \textbf{[T+T+T]} model is unable to focus on action words properly, instead it displays different sentence hierarchical patterns. 
As we suggested in the main text, intuitively a human-interpretable cooking-specific sentence encoder should be able to focus on the action words in written step-by-step instructions. For example, in Fig.~\ref{fig:sentence_tree_1_a}, \textbf{[T+T+L]} properly emphasizes on the word ``marinate'' by placing it closest to the root, while \textbf{[T+T+T]} places it at the deepest leaf in Fig.~\ref{fig:sentence_tree_2_a}. Similar observations about different hierarchical structures of other sentences encoded by two models can be seen in Fig.~\ref{fig:sentence_tree_1}(bcd) and Fig.~\ref{fig:sentence_tree_2}(bcd).
These evidences further support our hypothesis stated in the main paper: because sentences in a recipe are usually written in chronological order, trying to learn an instruction-level hierarchy actually hurts the ability to model the actions described in recipe instructions. In all of our experiments, the \textbf{[T+T+L]} model achieves the balance of emphasizing the importance of both ingredients and cooking actions.

\end{appendices}

\end{document}